%% file: 00_KDD_main.tex
\documentclass{article}

\usepackage{arxiv}

\usepackage[utf8]{inputenc} 
\usepackage[T1]{fontenc}    
\usepackage{hyperref}       
\usepackage{url}            
\usepackage{booktabs}       
\usepackage{amsfonts}       
\usepackage{nicefrac}       
\usepackage{microtype}      
\usepackage{lipsum}		
\usepackage{graphicx}
\usepackage{natbib}
\usepackage{doi}

\usepackage{courier}  
\usepackage{natbib}  
\usepackage{caption} 
\usepackage[ruled, vlined]{algorithm2e}
\usepackage{array}
\usepackage{amsmath}
\usepackage{bm}
\usepackage{booktabs}

\usepackage{enumitem}
\usepackage{graphicx}
\usepackage{multirow}
\usepackage{natbib}
\usepackage{soul}
\usepackage{xcolor}
\usepackage{caption}
\usepackage{subcaption}
\usepackage{mathtools}

\title{Experimentation Platforms Meet Reinforcement Learning: 
Bayesian Sequential Decision-Making for Continuous Monitoring}

\date{} 					

\author{%
  Runzhe Wan$^{*}$ \;\; Yu Liu$^{*}$ \;\; James McQueen \;\; Doug Hains \;\; Rui Song \\
  Amazon\\
  \texttt{\{wrunzhe, yluz, jmcq, dhains, ruisong\}@amazon.com} \\
}

\renewcommand{\headeright}{}
\renewcommand{\undertitle}{}

\renewcommand{\shorttitle}{
RL for  Early Termianting Experiments
}

\hypersetup{
pdftitle={A template for the arxiv style},
pdfsubject={q-bio.NC, q-bio.QM},
pdfauthor={David S.~Hippocampus, Elias D.~Striatum},
pdfkeywords={First keyword, Second keyword, More},
}

\usepackage[ruled,vlined]{algorithm2e}

\usepackage{natbib}
\setcitestyle{authoryear,round,citesep={;},aysep={,},yysep={;}}

\definecolor{mydarkblue}{rgb}{0,0.08,0.45}
\hypersetup{ %
colorlinks=true,
linkcolor=mydarkblue,
citecolor=mydarkblue,
filecolor=mydarkblue,
urlcolor=mydarkblue,
}
\renewcommand{\cite}[1]{\citep{#1}}

\input{Mis/new_commands}
\usepackage[ruled,vlined]{algorithm2e}
\usepackage{amsthm}
\usepackage{tikz}
\usepackage[edges]{forest}
\usepackage{mathtools}
\usepackage{stfloats}
\usepackage{array, makecell}
\usepackage{amsmath,bm}

\newcommand{\name}[1]{\texttt{#1}}

\let\svthefootnote\thefootnote
\newcommand\freefootnote[1]{%
  \let\thefootnote\relax%
  \footnotetext{#1}%
  \let\thefootnote\svthefootnote%
}

\newcommand\blfootnote[1]{%
  \begingroup
  \renewcommand\thefootnote{}\footnote{#1}%
  \addtocounter{footnote}{-1}%
  \endgroup
}

\begin{document}



\maketitle

\input{0_abstract}

\blfootnote{* Equal contribution}


\input{0_Intro}

\input{1_Related_Work}

\input{1_Setup}
\input{1_Preliminary}

\input{2_Formulation}

\input{2_formulation_discussion}

\input{3_Learning}

\input{4_Numerical}

\input{5_Conclusion}
\bibliographystyle{ACM-Reference-Format}
\bibliography{00_KDD_main}
\input{6_Appendix}

\end{document}

%% file: Mis/new_commands.tex
\newcommand{\vx}{\mathbf{x}}

\newcommand{\vthe}{\boldsymbol{\theta}}

\newcommand{\I}{\mathbb{I}}

\usepackage{soul}
\usepackage{xcolor}

\newcommand{\Mean}{{\mathbb{E}}}

\newcommand{\prob}{\mathbb{P}}

\usepackage{amsthm}
\newtheorem{thm}{Theorem}

\newtheorem{lemma}[thm]{Lemma}

\newtheorem{remark}{Remark}

\DeclareMathOperator*{\argmax}{arg\;max}

%% file: 0_abstract.tex
\begin{abstract}
With the growing needs of online A/B testing to support the innovation in industry, the opportunity cost of running an experiment becomes non-negligible. 
Therefore, there is an increasing demand for an efficient continuous monitoring service that allows early stopping when appropriate. 
Classic statistical methods focus on hypothesis testing and are mostly developed for traditional high-stake problems such as clinical trials, while experiments at online service companies  typically have very different features and focuses. 
Motivated by the real needs, in this paper, we introduce a novel framework that we developed in Amazon to maximize customer experience and control opportunity cost. 
We formulate the problem as a Bayesian optimal sequential decision making problem that has a unified utility function. 
We discuss extensively practical design choices and considerations. 
We further introduce how to solve the optimal decision rule via Reinforcement Learning and scale the solution. 
We show the effectiveness of this novel approach compared with existing methods via a large-scale meta-analysis on experiments in Amazon. 
\end{abstract}

%% file: 0_Intro.tex
\section{Introduction}\label{sec:intro}

Online A/B testing has become a common practice in online service companies (e.g., Amazon, Google, Netflix, etc.) to evaluate the customer impact of new features (treatments) in comparison to the old ones (control)~\cite{gupta2019top}. 
Unlike traditional randomized-controlled trials (RCT), in online A/B tests the experimental population is not known a priori and  \textit{convenience sampling} is often employed. If a given customer interacts with the feature being experimented on (e.g., streams a video, makes a search query, etc) they are shown either the treatment or control and are part of the experiment. As a result, the sample size of an online experiment using convenience sampling (as is the practice at Amazon) is a function of its duration and the longer an experiment is run the more samples are collected and the higher power we get. 

A/B experiments are typically conducted using \textit{fixed-horizon hypothesis testing}, where the duration of an experiment is predetermined in order to collect enough samples to achieve some desired statistical power (e.g., 80\%)~\cite{kadam2010sample, richardson2022bayesian}. 
However, running experiments has a non-zero cost. Longer experiments can slow down the innovation cycle and consume (potentially costly) hardware and human resources. The impact of the experiment on the customers is also a real concern. For example, a treatment with negative impact should be terminated early to reduce customer exposure to a negative experience. Similarly, positive treatments should be launched as soon as possible to maximize the customer benefit. In contrast to the fixed-horizon approach, methods like \textit{continuous monitoring} (also known as \textit{early termination}) allow us to update guidance on experiment duration while the experiment is still running.





Despite the extensive research on continuous monitoring (see Section \ref{sec:related_work} for a review), to the best of our knowledge, almost all of them are from the \textit{hypothesis testing} perspective, which focuses on the decision accuracy.  
The primary objectives are typically the false positive rate (FDR), type-I error or type-II error. 
This is probably due to the reason that most of these methods are from high-stake applications such as clinical trials, and the hypothesis testing viewpoint is known as conservative. 
In contrast, many online experimentation platforms have a different risk-opportunity profile and the properties of data are also very unique. 
For example, in online e-commerce companies, tens of thousands of experiments are run per year. 
Most of these experiments are not high-risk;  meanwhile, the majority of them cannot achieve statistical significance with desired powers, since the noise level of responses is high (given that the whole system is very complicated with numerous moving parts). In addition, online experiment platform are designed to encourage the exploration of novel ideas, but on the flip side many of them will not yield positive or significant results. The opportunity cost of conducting the experiment is also high due to the sibling teams are waiting on the same resources. 
For example, sibling teams that perform experiments on the same widget or impact the same customer group should run their experiments sequentially. During any active experiment, the other teams may need to wait their turn to access the same resources.
Therefore, finding an appropriate balance between customer benefits and opportunity costs is crucial for continuous monitoring A/B testing in online experimentation platforms. 
Classic methods that focus on the type-I error and power turn out to be too conservative, as we will show empirically.

\input{fig_illustration}

We illustrate our motivations using real experiments conducted in Amazon. 
Figure~\ref{image:histogram} displays the \textit{ex-post} measured powers of a large number of experiments. 
The histogram shows that there is a large number of experiments with very low powers, which we can terminate earlier since they are not able to achieve the significance level (with high probability). 
There is also a large number of experiments with very high powers, for which we can terminate earlier as well, since the evidence becomes strong in a short time window. 
In Figure~\ref{image:trajectory}, we further zoom in by looking at the trajectories of z-statistics of a few representative experiments. 
Z-statistics are calculated using a fixed-horizon two-sided z-test with a significance level of $0.05$ \cite{welch1947generalization}. 
Experiment 2 shows consistently positive z-statistics (i.e. treatment is consistently better than control), 
experiment 3 shows consistently negative z-statistics, 
and the z-statistics of experiment 1 are close to zero. 
It is intuitive to consider terminating these experiments earlier. 
In conclusion, applying earlier termination can be beneficial a lot in online A/B testing, in terms of efficiency and resource allocation. 

Motivated by these real needs, 
we are concerned with the following question: 

\textit{
How can we efficiently and appropriately early terminate an experiment, in a way that comprehensively considers and balances different sources of costs and benefits to maximize the customer utility? 
}




\textbf{Contribution.} 
Motivated by the real needs we observed for online experiments, we propose a continuous monitoring service that can provide early termination recommendations when appropriate. 
We propose to formulate the problem as a Bayesian sequential decision making (SDM)  problem~\cite{berger2013statistical}. 
We design a concrete yet general framework that appropriately formulates the online A/B testing into SDM. 
The objective function is the expected cumulative utility from an online A/B experiment, where the decision accuracy, the opportunity costs and the experimentation impacts are all taken into consideration. 
We present a list of questions frequently asked in practice to further elucidate the framework, and also discuss a few extensions to accommodate different real needs.
We also discuss how to solve the problem via RL and how to scale our solution by leveraging the contextual RL technique. 
Finally, we present a large-scale meta-analysis on a large number of past experiments at Amazon. 
The results clearly demonstrate the advantage of RL, as an effective and practical solution for early stopping online experiments.  
To the best of our knowledge, this paper is the \textit{first} systematic study of RL-based approach for continuous monitoring. 
The paper aims to provide a detailed guidance on how to apply these techniques in industrial settings.

\textbf{Outline. } We organize the paper as follows. 
In Section \ref{sec:related_work}, we introduce the related works. 
Section~\ref{sec:setup} gives the problem setting and assumptions which are consistent with Amazon's practice. 
In Section~\ref{sec:NMDP}, we provide some preliminaries on SDM. 
Section~\ref{sec:formulation} provides the details of our approach, which is supported by our meta-analysis results in Section \ref{sec:experiment}. 
Section \ref{sec:conclusion} concludes this paper.

%% file: fig_illustration.tex
\begin{figure}
    \centering
    \begin{minipage}{0.45\linewidth}
        \centering
        \includegraphics[width=\linewidth, height = 6cm]{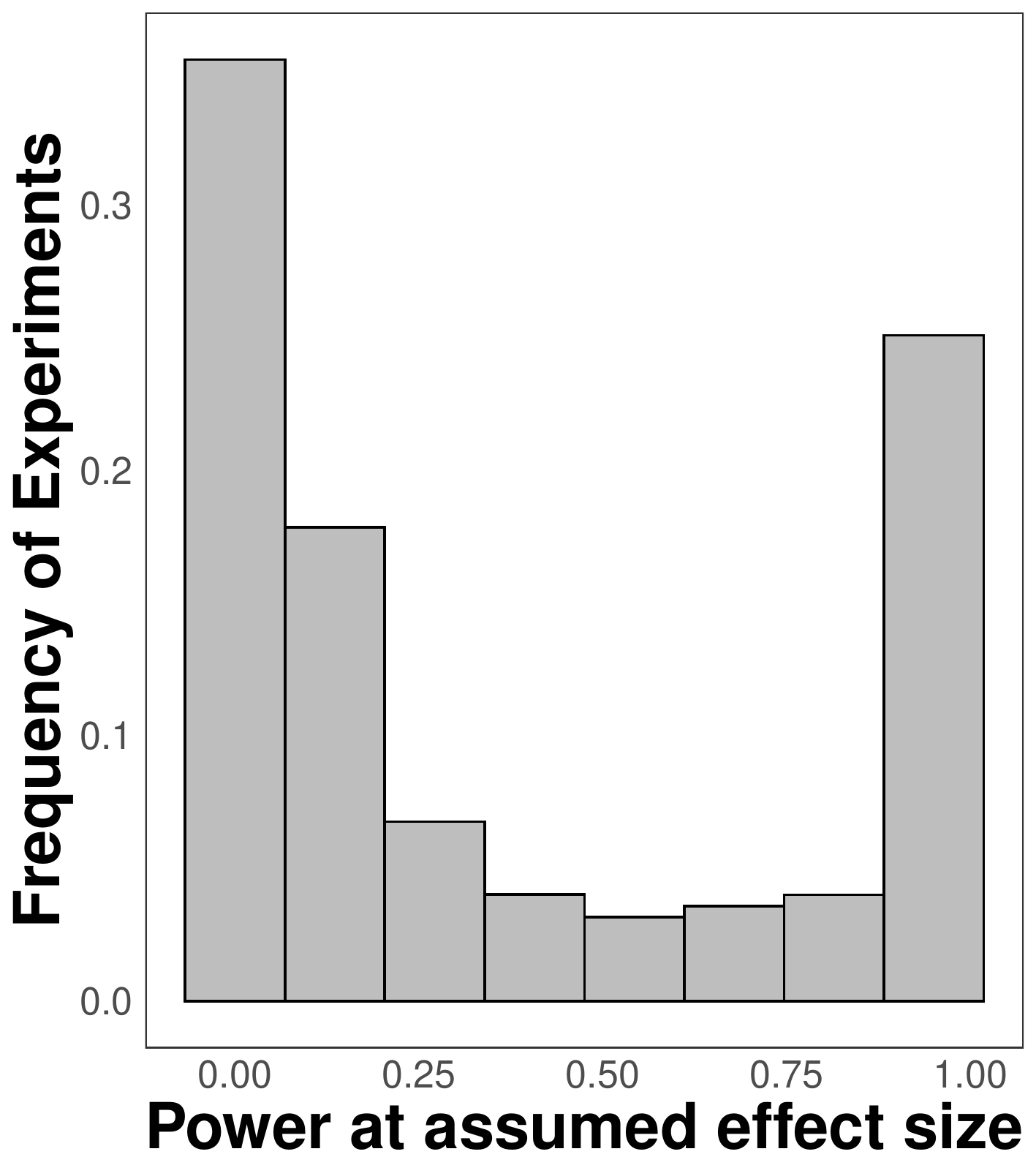} 
        \caption{Histogram of the frequency of the power values (at a given assumed effect size) at the end of the experiments. 
        }
    \label{image:histogram}    
    \end{minipage}
    \hfill
    \begin{minipage}{0.52\linewidth}
        \centering
        \includegraphics[width=0.8\linewidth, height = 6cm]{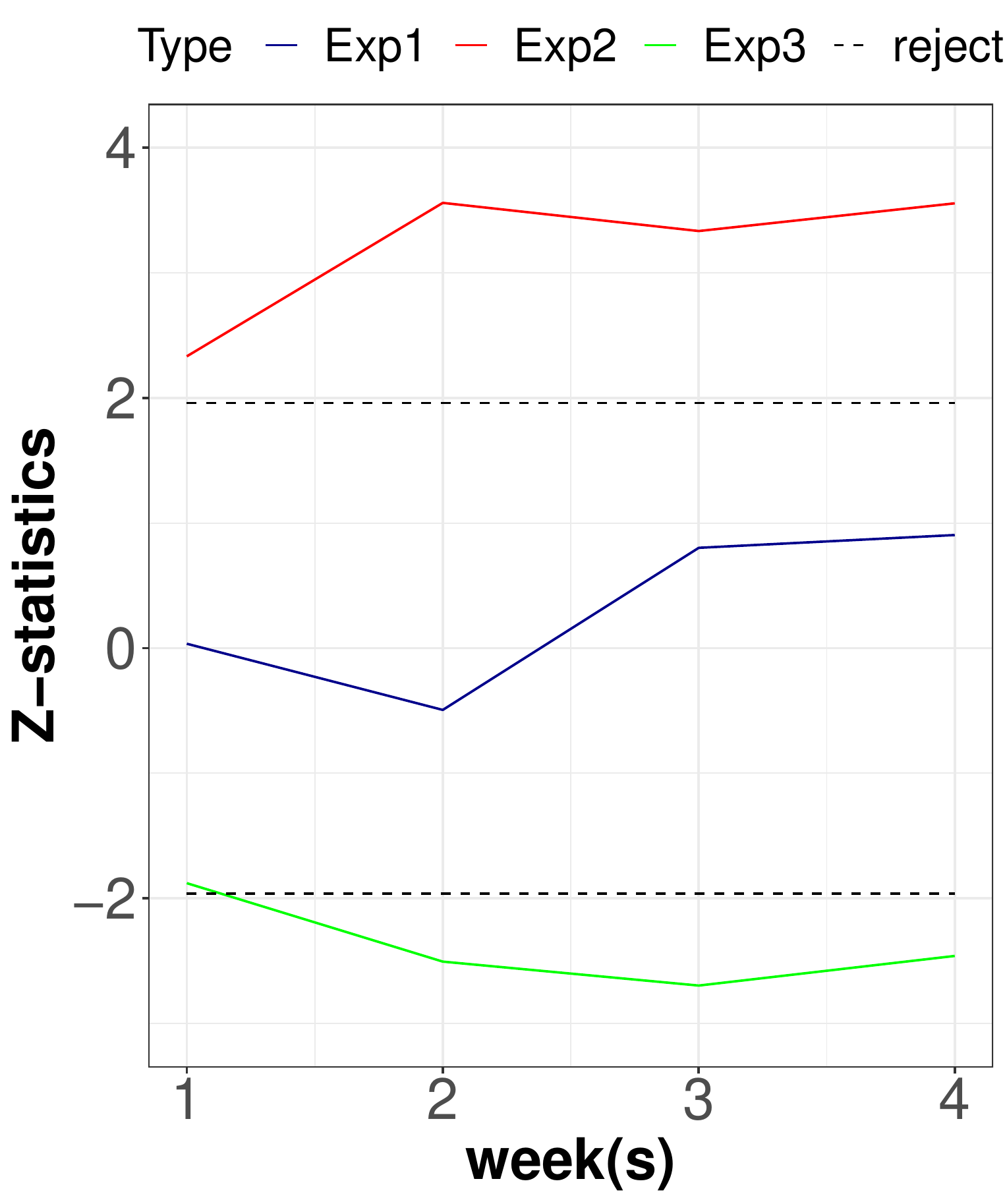} 
        \caption{Trajectories of z-statistics for three actual experiments. 
        The dotted lines indicate the rejection boundary.}
        \label{image:trajectory}
    \end{minipage}
\end{figure}

%% file: 1_Related_Work.tex
\section{Related Work}\label{sec:related_work}

\textbf{Early termination. }
In the statistics literature, there is a long history on studying how to detect the true treatment effect as quickly as possible or give up sooner if there is no hope to detect anything. 
The alpha-spending function approach is proposed \cite{demets1994interim, gordon1983discrete} to control the type-I error rate across the interim tests. 
In Bayesian hypothesis testing, the sequential Bayes factor (SBF) procedure, which is similar to the sequential testing proposed in~\cite{wald1948optimum}, is widely used~\cite{jeffreys1961theory, schonbrodt2017sequential}. 
~\cite{deng2016continuous} shows the validity of Bayes Factor and proposed a stopping rule based on it, which controls the false discovery rate (FDR). 
Always valid inference~\cite{johari2017peeking, johari2022always} is proposed as a user-friendly framework to control the type-I error, by converting any existing sequential testing procedure into a sequence of p-values. 
A few methods (such as the conditional power or the predictive power approach) that predict the outcome of the experiment so as to early terminate are also designed \cite{kundu2021conditional}. 
We provide more details of these methods in Appendix~\ref{appendix:baseline}. 
However, all of them focus on conventional statistical accuracy metrics, and hence the objectives are different from ours. 



\textbf{RL for experiment design. }
RL has been applied to many real domains including healthcare \citep{murphy2001marginal}, robotics \citep{andrychowicz2020learning}, epidemiology \cite{wan2021multi} and autonomous driving \citep{sallab2017deep}. 
In recent years, RL has been also  successfully applied to sequential experiment design. 
\cite{muller2007simulation} proposes a few simulation-based methods for maximizing the expected utility, which however is not computationally feasible with continuous spaces. 
RL techniques such as  approximate DP \citep{huan2016sequential} and 
policy gradient \citep{shen2021bayesian} are later applied to obtain an approximate solution. 
This problem is recently extended to even more challenging cases, where an explicit model is not available but sampling from the distribution is feasible \citep{ivanova2021implicit, foster2021deep}. 
\cite{tec2022comparative} compared RL with simulation-based methods and discussed its potential applications in clinic trials. 
Overall, few attention is focused on the early stopping problem and it is only sometimes used as a toy example. 
To the best of our knowledge, this paper is the \textit{first} systematic study of RL-based approach for early stopping. 
We advance both methodology development and industrial applications on this topic. 


%% file: 1_Setup.tex
\section{Problem Setup}\label{sec:setup}
In a typical online A/B test, customers may visit the site multiple times, and when customers visit  can't be controlled. 
Every visit generates an outcome of interest. 
However, typically an online site is  very complicated and a new feature is incremental. 
Therefore, to reduce the noise, it is standard to only focus on those outcomes after a costumer is \textit{triggered} based on some experiment-specific trigger mechanism definition (e.g., after she saw the related features) . 
After that, we call the customer has entered the experiment \citep{richardson2022bayesian}. 
Besides, unlike traditional testing settings where there are a small number of i.i.d. units, online experiments may have millions of units and are non-stationarity. 
It is not practice to make a decision with every new data point. 
Therefore, we follow this tradition to aggregate users' responses on a weekly basis. 
 Other choices of the time unit are also possible, depending on the applications.





\textbf{Notations.}  Let $T$ be the horizon length, which represents the maximum allowed duration (number of weeks) for an experiment. 
Denote $N_{k, g}$ as the number of customers who are triggered at week $k$ in group $g \in \{Tr, C\}$. 
In this paper, $Tr$ denotes the treatment group and $C$ denotes the  control group. 
Denote the number of customers and the set of those customers who are triggered between week $t$ and $l$ as $N_{t:l, g}$ and $\mathcal{I}_{t:l, g}$, respectively, where $N_{t:l, g} = \sum_{k=t}^{l} N_{k, g}$.
For a customer $i$, denote $W_{i, k}$ as her (aggregated) outcome in week $k$, and $W_{i, t:l} = \sum_{k=t}^{l} W_{i, k}$ as her cumulative outcome between week $t$ and $l$.
A customer begins to generate outcome after being triggered. 

\textbf{Assumptions on stationary treatment effects}. 
For ease of exposition of our main idea, we assume the treatment effects are stationary over time. 
Formally, we assume 
 $E[W_{i, t}] = \mu_g, \forall i \in \mathcal{I}_{1:k, g}$ and $\forall t \in\{1, \ldots, T\}$, where $\mu_g$ is the population mean of weekly aggregated response for group $g \in \{Tr, C\}$. 
 We will discuss how to relax this assumption. 
  We further define the per-customer per-week treatment effect as $\delta = \mu_{Tr}- \mu_{C}$.







\textbf{Earlier Termination.} 
With a fixed-horizon procedure, we always wait until time $T$ and use all data to make the decision. 
Since we observe data sequentially over time, we can  consider terminating earlier when appropriate. 
Specifically, we can make a decision at each time point $t$, such as whether to continue the experiment, launch the new feature or not, based on all available data. 

%% file: 1_Preliminary.tex
\section{Preliminaries}\label{sec:preliminaries}
We first introduce a general formulation of the Sequential Decision Making (SDM) problem. 
We focus on the finite-horizon undiscounted non-stationary setup, which is most relevant to our problem. 
We start with the (potentially) non-Markovian setup in  Section \ref{sec:NMDP}, which is most general and hence flexible. 
In Section \ref{sec:MDP}, we discuss the Markovian variant, and introduce the belief states for the Bayesian perspective. 

\subsection{Sequential Decision Making}\label{sec:NMDP}
Let the horizon be $T$. 
A (potentially) \textit{non-Markov decision process (NMDP)} is defined by a series of 
observation space $\mathcal{Y}_t$, 
action space $\mathcal{A}_t$, 
transition kernel $\mathcal{P}_t$, 
and reward kernel $\mathcal{R}_t$ for every $t = 1, \dots, T$. 
Let $\{(Y_t, A_t, R_t)\}_{t\ge 1}$ denote a trajectory generated from the NMDP model, where $(Y_t,A_t,R_t)$ denotes the observation-action-reward triplet at time $t$. 
Denote $\mathcal{H}_t = (Y_1,A_1,R_1, \dots ,Y_{t-1},A_{t-1},R_{t-1}, Y_t)$ as the historical information available until time $t$. 
We denote all unknown parameters of the system by $\vthe$. 
In the Bayesian setting, we assume a prior over $\vthe$ as $\rho(\vthe)$. 
To simplify the presentation, we assume the probability spaces are continuous. 
The transition kernel $\mathcal{P}_t(Y_t| \mathcal{H}_{t-1}, A_{t-1}; \vthe)$ gives the probability density of observing $Y_t$ by taking action $A_{t-1}$ given the history $H_{t-1}$, and similarly 
$\mathcal{R}_t$ generates the reward. 
A (deterministic) history-dependent \textit{policy (a.k.a. decision rule)} $\pi = (\pi_1, \dots, \pi_T)$ is a series of functions, where $\pi_t$ map the history to an available action $A_t = \pi_t(\mathcal{H}_t) \in \mathcal{A}_t$. 

A trajectory following policy $\pi$ is generated as follows. 
The agent starts from an observation $Y_1$. 
At each time point $t \ge 1$, the agent selects an action $A_{t} = \pi_t(\mathcal{H}_{t})$, then receives a random reward $R_{t} \sim \mathcal{R}_{t}(\cdot \mid \mathcal{H}_{t}, A_{t} ; \vthe)$, and finally observes the next observation $Y_{t+1} \sim \mathcal{P}_{t}(\cdot \mid  \mathcal{H}_{t}, A_{t}; \vthe)$. 
For a policy $\pi$, its history value function (V-function) and history-action value function (Q-function)~\cite{sutton2018reinforcement} are defined as 
\vspace{-0.1cm}
\begin{eqnarray*}
	V^{\pi}_{t'}(h)=\Mean^{\pi} \Big( \sum_{t=t'}^{T} R_{t}|\mathcal{H}_{t}=h \Big), Q^{\pi}_{t'}(a,h)=  \Mean^{\pi} \Big(\sum_{t=t'}^{T} R_{t}|A_{t}=a, \mathcal{H}_{t}=h \Big),
\end{eqnarray*}
where the expectation $\Mean^{\pi}$ is defined by assuming the system follows the policy $\pi$. 
More specifically, the expectation is taken over trajectories with $A_t = \pi_t(\mathcal{H}_t), \forall t$,  and the other dynamics following the underlying MDP model.

The objective is to find an optimal policies $\pi^* = (\pi_1^*, \dots, \pi_T^*)$ following which we can maximize the expected cumulative reward (i.e., the utility). 
It can be defined as $\pi^* \in \argmax_{\pi} V^{\pi}_{t'}(h), \forall t', \forall h.$ 
The optimal Q-function is defined as:
\begin{align*}
Q^*_T(a, h) &= \Mean \Big(R_{T}|A_T =a, \mathcal{H}_{T}=h \Big) \\
Q^*_{t}(a, h) &= \Mean\big[R_t + \max_{a' \in \mathcal{A}_{t+1}} Q^*_{t+1}(a', \mathcal{H}_{t+1})  |A_t =a, \mathcal{H}_{t}=h  \big] , \forall t < T
\end{align*}
 we have $\pi_t^*(h) = \argmax_{a \in \mathcal{A}_t}Q^{*}_{t}(a, h)$, for any $h$ and $t$.   




\subsection{Markov Decision Process}\label{sec:MDP}

When additional structures are satisfied, we can consider a more efficient decision process model, the Markov Decision Process \citep[MDP,][]{puterman2014markov}. 
The (Bayesian) MDP is particularly relevant when disucssing the model that we are working with (see Section \ref{sec:N-N}). 

Specifically, 
suppose we can construct some observable so-called state variable $S_t$ at every time point $t$.   
The key assumption of MDPs is the  Markov assumption (with slight overload of notations)
\begin{align*}
	&\prob(S_{t+1}|\{Y_{j},A_{j},R_{j}\}_{1\le j\le t}  ; \vthe)=\mathcal{P}_t(S_{t+1} \mid A_{t},S_{t} ; \vthe),\\
	&\prob(R_{t}|S_{t},A_{t},\{Y_{j},A_{j},R_{j}\}_{1 \le j< t}  ; \vthe) = \mathcal{R}_t(R_t \mid A_{t},S_{t}  ; \vthe). 
\end{align*}
The assumption requires that there is no information useful for predicting the future being omitted from the state vector. 
In other words, roughly speaking, the state vector is a sufficient statistic of the history. 
Under an MDP, the optimal policy can only depend on the current state $S_t$ instead of the whole history $\mathcal{H}_t$, so do the Q- and V-function for this class of policies.






\textbf{Bayesian MDP and belief states. }
In the Bayesian setup, the unknown parameter $\vthe$ is a random variable, which becomes a confounding variable and typically invalidates the Markov property. 
The model becomes a so-called partially observable MDP (POMDP). 
Fortunately, one can adopt a classic technique to transform such a Bayesian problem as an MDP, by constructing the so-called \textit{belief state}. 
Specifically, we can include the posterior of $\vthe$, $P(\vthe | \mathcal{H}_t)$, as a component of the state $S_t$. 
Then it is easy to check that the problem becomes an MDP again \citep{vlassis2012bayesian}. 
When the posterior belongs to a parametric class, it is sufficient to include the parameters of the posterior (i.e., the sufficient statistics that capture all historical and prior information). 
One can also show that, the transition and reward function can be equivalently written as being based on the marginalized distribution over the posterior  $P(\theta | \mathcal{H}_t)$. 


%% file: 2_Formulation.tex
\section{Early termination with RL}\label{sec:formulation}
We present our proposed framework in this section. 
We first discuss the choice of the overall RL approach in Section \ref{sec:RL}. 
We then lay down the model formulation in Section \ref{sec:MDP formulation}, 
and next discuss this formulation and its extensions in Section \ref{sec:discussion_model}, 
We intentionally make the formulation general, so that it can be adapted to different real needs. 
Finally, we introduce how to solve an optimal policy via RL and how to scale the solution in Section \ref{sec:optimization}.

\subsection{Policy Optimization with Model-based Bayesian RL}\label{sec:RL}

Reinforcement Learning  \citep[RL,][]{sutton2018reinforcement} solves the optimal policy $\pi$ by aggregating information from data (trajectories). 
There are multiple ways to classify RL methods. 

First of all, RL methods can be dichotomized as \textit{model-based} v.s. \textit{model-free}. 
The model-based approach requires knowledge of (or assumption on) the transition and reward kernels $\mathcal{P}$ and  $\mathcal{R}$, 
while the model-free approach only assumes the environment is an NMDP and directly learns from the trajectories. 
The former approach is typically more efficient, at the cost of needing knowledge (assumptions) on the model structure. 
In our problem, 
there are a lot of problem-specific structures we can utilize (e.g., the experiment terminates when we take the corresponding actions; see Section \ref{sec:MDP formulation} for more details). 
Moreover, we observe that different experiments differ significantly. 
A model-free approach needs to learn one policy from them all (without model structure) and hence would be sub-optimal. 
Therefore, it is natural to adopt the model-based approach. 

Second, RL approaches can also be divided as \textit{online} and \textit{offline}: the former continuously interacts with the real environment, while the latter does not. 
In our applications, it is not feasible to trial and error with every new experiment. 
Therefore, we utilize historical information to build a model for the new experiment, use the model as a simulator and solve it via online RL algorithms. 
The overall framework is offline. 

Finally, we adopt a Bayesian approach, which allows us to utilize valuable prior information and also formulate an optimization problem in a natural way. 
Such a choice is classic and closely related to the vast literature on Bayesian decision theory \cite{gelman1995bayesian} and Bayesian experiment design \cite{chaloner1995bayesian}.

\subsection{SDM Problem Formulation}\label{sec:MDP formulation}

\input{diagram_SDM}

In the following subsections, we define the components of our model. 
We exposit under the most general non-Markovian scenario.  
One example about our application of this framework at Amazon will be given in Section \ref{sec:N-N}, where we consider a special Markovian case. 
Figure~\ref{fig:dpg} provides an illustration.


We consider a \textit{finite-horizon} setup, where at every decision point $t = 1, \dots, T$, we will make a decision. 
Note that, when the experiment has already been terminated, any action below will not make any difference. 
Alternatively, one can also consider an \textit{indefinite-horizon} setup, based on which we observed similar performance. 
Recall that the $t$-th decision point is defined as the \textit{end} of the $t$-th time period.


 



\subsubsection{Unknown Parameters}
The unknown parameter vector $\vthe$ contains at least the mean response of customers in the two groups, i.e., $\mu_{Tr}$ and $\mu_{C}$.
$\vthe$ can also contain other unknown quantities, such as the  sample size distribution and the variance/covariance terms, etc., depending on design choice. 
In these cases, we only need to include the related data in the observation vector and modify the components to be introduced accordingly. 

In practice, there is a rich literature on sample size prediction ~\cite{burnham1978estimation,anderson2016library}, and with the model proposed in ~\cite{richardson2022bayesian} we observed a high accuracy and low uncertainty. 
Therefore, to avoid unnecessary complexity and for ease of exposition, we assume the sample sizes in every week are known \textit{a priori}. 
Similarly, we assume the variances of outcomes are known, which in practice are typically plugged in with the sample variances. 
The framework can be naturally extended by treating the uncertainty and estimation of these components as part of the transition dynamic. 





We assume we have a prior distribution $\rho(\vthe)$. 
We use $\Mean_{\rho}$ to denote $\Mean_{\vthe \sim \rho(\vthe)}$, i.e.,  taking expectation over this prior.

\subsubsection{Observations}
In week $t$, if the experiment remains continuing, the observation $Y_t$ contains the outcomes of all customers triggered before the end of that week, i.e.,  $\{W_{i,t}\}_{i \in \mathcal{I}_{1:t, Tr}}$ and 
$\{W_{i,t}\}_{i \in \mathcal{I}_{1:t, C}}$. 
$Y_t$ may also include other observed variables useful for estimating $\vthe$. 
If the experiment has been terminated, we denote $Y_t = \emptyset$. 
Because in our case the observations will become part of the history, we discuss the observation generation model in Section \ref{sec:state}, where we focus on the transition kernel.


\subsubsection{Actions}\label{sec:action} 

From a high level, there are three categories of decisions we can make at every decision point, including (i) when to stop the experiment, (ii) if stopped, whether to launch the new feature, and (iii) what statistical statements to make (e.g., the p-value of a given metric). 
There may be many variants of the action space. 
In this paper, we will focus on the following three options that we are working with: 
\begin{itemize}
    \item a = 0: keep running; 
    \item a = 1: stop and launch; 
    \item a = 2: stop and no launch. 
\end{itemize}



\textbf{Action spaces (feasible action set). }
It is natural to set $\mathcal{A}_t = \{0, 1, 2\}, \forall t < T$, and $\mathcal{A}_T = \{1, 2\}$. 
Let $I_t$ be the indicator for the event $\bigcup_{t' < t} \{A_{t'} \neq 0\}$, i.e., the experiment has been terminated before time $t$. 
We note that, when $I_t = 1$, any choice of actions does not make difference, as the experiment has terminated. 




\subsubsection{Objective and Reward}\label{sec:objective_reward} 
In this section, we give our optimization objective. 
Specifically, for any trajectory $\{(A_t, Y_t)\}_{t \le T}$ and any value of the unknown parameter $\vthe$, 
we define an utility function  and aim to solve a policy that maximizes the expected utility. 
The utility should include (at least) the following components:
\begin{enumerate}
    \item $c$: we assume there is a fixed and pre-specified \textit{weekly cost for running the experiment}. This may include (with some potential overlap): 
    The \textit{personel and harware cost} to support this specific experiment, 
    the \textit{opportunity cost for the experimentation platform}, 
    and the  \textit{opportunity cost for the experimenters}. 
    \item $c_h$: The \textit{hurdle cost}, which is the the cost needed to launch (e.g., the implementing in production) new feature. 
    \item The \textit{customers impact from making the launch recommendation} on weeks $t$. This term concerns the decision accuracy, i.e., we would like to launch those new features that indeed have positive customer impacts. 
    \item The \textit{customers impact on the treatment group from the experiment itself}.
    For example, if the treatment has clear negative impacts on customers, it should be terminated earlier. 
\end{enumerate}
We assume all these components share the same unit $D$, which can be monetary or something else that measures customer impacts.  




\textbf{Utility of an experiment.} 
Notice that, since ATE is a relative value, our utility/reward definitions below are also relative to the outcomes under the control.  
The utility can then be defined as
\begin{align}\label{eqn:extended_utility_old}
&u(\{(A_t, Y_t)\}_{t \le T}, \vthe) \nonumber\\
    &= 
    \underbrace{-c \times \sum_{t=1}^T\I(I_t = 0, A_t = 0) }_\text{Opportunity Cost} 
    + 
    \underbrace{
    \sum_{t=1}^T
    \big( \delta \times N_{1:t+1, Tr}
    \big)
    \I(I_t = 0, A_t = 0)
    }_{\mathclap{\substack{\text{Impact during the experiment}\\ \text {(relative to control)}}}} \nonumber\\
    &+ 
    \underbrace{
    \sum_{t=1}^T 
      (u_2(\delta, H, T, t) -c_h)\I(I_t = 0, A_t = 1), 
    }_\text{Launch impact (relative to control)}
\end{align}
where $N_{1:l, g}$ denotes the cumulative number of triggered customers in group $g \in \{\text{Tr, C}\}$ and recall $\delta$ is the per-customer per-week average treatment effect  defined in Section~\ref{sec:setup}. 
The first term represents the opportunity cost incurred for conducting the experiment until termination. 
The second term represents the impact on the treatment group during the experiment. 
If the treatment has a clearly negative impact, the second term will encourage we to terminate the experiment earlier. 
In the third term, $u_2(\delta, H, T, t)$ represents the impact from launching the treatment feature on the whole population at week $t$, given  a pre-specified time horizon $H$. 
Typically, we assume a one-year time horizon ($H = 52$ weeks) and linear extrapolation, with all customers triggered in the experiment period regarded as the target population. 
This gives an example of the launch impact as $u_2(\delta, H, T, t) = \delta \times (H + T - t) \times (N_{1:T, Tr} + N_{1:T, C}) $. 
We note that, the third term has an decreasing relationship with the week we recommend launch. 
Thus, the third term encourages launching a good feature earlier. 

We emphasize that, Equation~\eqref{eqn:extended_utility_old} does not include the impact and cost in week $1$, since they are constants that our policy does not have control on (any experiment needs to run for at least one week). 
Besides, the first two terms correspond to the impact on week $t+1$ from the decision made at the end of week $t$ (i.e., at the $t$-th decision point). 
These two terms are always $0$ at $t=T$. 
This is because the experiment will end no later than $T$, so $A_T \neq 0$ (recall our definition of the action spaces).

\textbf{Objective. }
Our objective is to solve a policy $\pi^*$ that maximizes 
\begin{align}\label{eqn:objective}
    \Mean^\pi_{\vthe \sim \rho(\vthe)}
    & u \Big(\{(A_t, Y_t)\}_{t \le T}, \vthe \mid Y_1 = y \Big)
\end{align} 
for any $y$. 
Here, the expectation is taken over the trajectories following $\pi$. 
In other words, at every time point, given the prior, the observed data, and some other known components needed for modeling the transition (such as the sample size prediction model), we aim to determine the optimal policy (decision rule) that maximizes the expected utility, in the Bayesian sense.



\textbf{Rewards. }
To apply RL algorithms, we decompose the objective into the per-round instant reward $R_t$ in a way that the objective is equal to $\Mean^\pi_{\rho} \sum_{t=1}^T R_t$. 
There exist multiple equivalent ways. 
We consider a natural way as
\begin{equation}
R_t =
\begin{cases}
     -c + \Mean_{\rho}\big[\delta \mid \mathcal{H}_t \big] * N_{1:t+1, Tr}, &\text{when}  \; A_t = 0 \; \text{and} \; I_t = 0, 
     \\
     \Mean_{\rho}\big[  \delta \mid \mathcal{H}_t \big] * N_{1:T}  (H + T - t) -c_h,
     &\text{when} \; A_t = 1 \; \text{and} \; I_t = 0, 
    
     \\
     0, & \text{when} \;\;A_t = 2  \; \text{and} \; I_t = 0, 
     \\
     0, & \text{when}\;\;I_t = 1, 
     \end{cases}
     \label{eq:reward}
\end{equation}
where 
$N_{1:T} = \sum_{g \in\{Tr, C\}} N_{1:T, g}$. 
When defining the impact from the treatment (in the first two cases), 
we replaced $\delta$ in \eqref{eqn:extended_utility_old} by $\Mean_{\rho}\big[\delta \mid \mathcal{H}_t \big]$. 
This is a common trick which gives an equivalent objective and reduces the randomness to make training easier. 


\subsubsection{Transition Kernel}\label{sec:state}
The last piece of the formulation is the transition model $p(Y_{t+1} \mid A_t, \mathcal{H}_t)$, i.e., the distribution of our observation in the next period given the history, our current action, and the prior. 
Based on the Bayes rule, we can first sample $\Tilde{\vthe}$ from $p(\vthe \mid A_t, \mathcal{H}_t) = p(\vthe \mid \mathcal{H}_t)$ and then sample from $p(Y_{t+1} \mid A_t, \mathcal{H}_t, \Tilde{\vthe})$. 
We will focus on discussing $p(Y_{t+1} \mid A_t, \mathcal{H}_t, \vthe)$, because $p(\vthe \mid \mathcal{H}_t)$ is essentially an Bayesian estimation problem and should have been studied extensively in an experimentation platform.

By definition, we have $p(Y_{t+1} = \emptyset \mid A_t = a, \mathcal{H}_t = h, \vthe) = 1$ whenever $I_t = 1$ or $a \in \{1,2\}$. 
In other words, the experiment has been terminated and there are no further observations. 
In all the other cases, this requires us to model the distribution of customer-level outcomes in the next time period, conditioned on all customer-level observations so far and the value of $\vthe$. 

If the observations in different weeks are independent, 
then sampling based on $p(Y_{t+1} \mid A_t = 0, \mathcal{H}_t, \vthe)$ would be straightforward: since $Y_{t+1}$ are independent from $\mathcal{H}_t$ conditioned on $\vthe$ (i.e., $p(Y_{t+1} \mid A_t = 0, \mathcal{H}_t, \vthe)$  = $p(Y_{t+1} \mid A_t = 0, \vthe)$ ), 
we only need to sample i.i.d. data points from the corresponding environment model with known parameters $\vthe$. 
Furthermore, by the central limit theorem, we can typically avoid the need of specifying a parametric model for each data point and can focus on the average outcome (see Section \ref{sec:N-N} for more details). 
In some other applications where the observations could be dependent over different weeks (e.g., the outcomes from the same customer can be correlated), typically we apply a repeated measurement model (see, e.g., \cite{lindsey1999models}). 

Since the choice of the outcome model crucially depends on the application, we will not give one detailed form in this section. 
In our case study in Section \ref{sec:N-N}, we present a model that we are working with  at Amazon. 







%% file: diagram_SDM.tex
\begin{figure} 
	\centering 
     \includegraphics[width=0.22\textwidth]{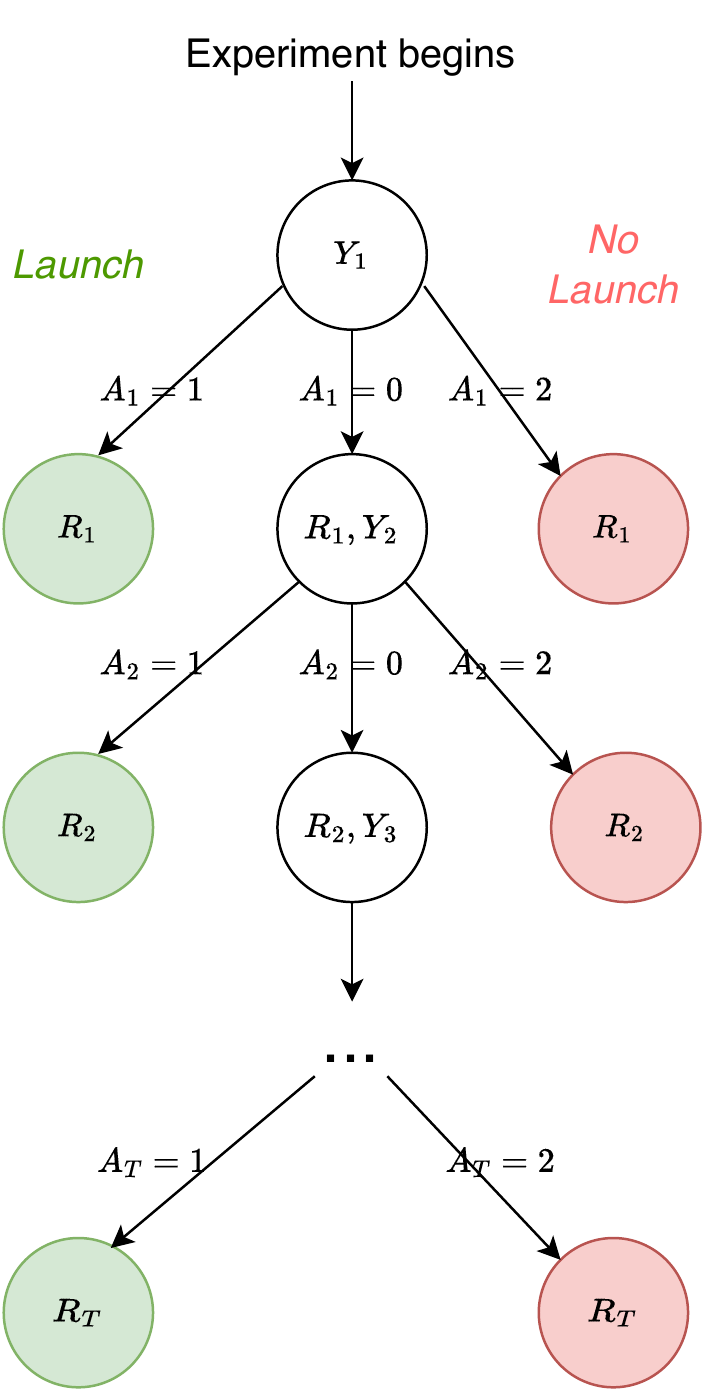} 
  \caption{Sequential Decision Making Process for early terminating experiments. Experiment starts at week $0$ and must continue for at least one week. 
  At each decision point $t$ (except for the last one), we can choose from "continue", "stop and launch", and "stop and no launch". 
  A reward $R_t$ will then be realized. 
  }\label{fig:dpg}
\end{figure}

%% file: 2_formulation_discussion.tex
\subsection{Discussion and Extensions of the Formulation}\label{sec:discussion_model}
In this section, we list a few questions frequently asked in our practice, to help elucidate the formulation and its natural extensions. 
We defer some more complicated extensions to the conclusion section. 

\textbf{Guardrails. }
Experiments may have certain safety guardrails on some metrics - once these guardrails are violated, the experiment will be terminated automatically. 
Such a mechanism can be naturally incorporated into our framework by modifying the transition function: once the constraints are violated, we set $I_t = 0$ automatically no matter what action the policy takes. 


\textbf{Statistical inference. }
Our formulation focuses on utility maximization, which is different from pure statistical inference and fits our applications better. 
However, in some cases, experimenters may still want some valid statistical statements along with the recommendation from the experimentation platform. 
We provide two approaches to add this important feature. 
First, we can modify the action space so as RL only determines whether or not we should terminate, and we slightly modify the transition and reward functions to \textit{automatically} make the launch decision based on the always valid p-values (AVP) \citep{johari2015always} when terminated. 
As such, our procedure satisfies the requirements in \citep{johari2015always} and hence the launch decision has a valid type-I error control. 
Such a decision rule has the highest utility among all rules that satisfy the type-I error bound. 
Alternatively, we can formulate a multi-objective problem, where the objective is the sum of the original utility and an indicator of making the wrong decision (multiplied by a parameter $\lambda$). 
This is essentially a Lagrange transformation. 
We can tune $\lambda$ until the error rate falls below a desired threshold, or share a list of policies under different values of $\lambda$ to provide users with choices.

\textbf{Multi-armed experiments. }
The extension to multi-armed experiments is straightforward, by duplicating (and slightly modifying) the treatment-related components in the state and action definitions. 

\textbf{ATE estimators. } 
The framework introduced in Section \ref{sec:MDP formulation} is general and not restricted to a specific ATE estimator. 
For example, one can apply covariate adjustment~\cite{lin2013agnostic} or model the heterogeneity~\cite{richardson2022bayesian} among customers via hierarchical modeling. 
One only needs to keep the related variables in the observation vector and modify the transition kernel accordingly. 







\textbf{Role of information gain}. 
The weekly cost $c$ clearly reflects the penalty for running long experiments. 
One natural question people frequently ask is how does the formulation encourage gaining information, which is central to an experiment. 
This may not be explicit, since our objective, as the \textit{expected} utility, seems not related to the \textit{uncertainty}. 
The answer is, this component has been naturally but implicitly considered in our objective. 

To see this clearly, we present an illustrative example below to show how does our objective encourages gaining information. 
we first introduce the fixed-horizon Bayesian optimal decision rule where a final decision $\pi_{1,t}(\mathcal{H}_{t}) = 2 - \I\big[  \Mean_{\rho}(\delta \mid \mathcal{H}_{t} ) \times H \times N > c_h \big]$ is made at $t$, (i.e., $\mathcal{A}_l =  \{0\}, \forall l < t)$. 
In other words, we launch when the posterior mean of the (annual) launch impact is higher than the hurdle cost. 
We consider the following example. 



\begin{lemma}
\label{lemma:reward-diff}
Assume two policies $\pi_1$ and $\pi_2$: 
$\pi_1$ waits until $t = t'$ to make a decision $A_t' \in \{1,2\}$ following the fixed-horizon Bayesian optimal decision rule; 
while $\pi_2$ waits until $t = t' + 1$ to do so. 
To simplify the notation, we assume $c_h = 0$. 
For any history $h$ at time $t'$, the difference of expected accumulative rewards (defined in~\eqref{eq:reward}) between the two policies is
\begin{align*}
	&\Mean_{\rho}^{\pi_2} \Big[\sum_{t=t'}^T R_t \mid \mathcal{H}_{t'} = h  \Big] - \Mean_{\rho}^{\pi_1} \Big[\sum_{t=t'}^T R_t \mid \mathcal{H}_{t'} = h \Big]\\
    &=
	\bigg\{\Mean_{\rho} \Big[  \Mean_{\rho}\big[\delta \mid \mathcal{H}_{t'+1} \big] \times \I[\pi_{2,t}(\mathcal{H}_{t'+1}) = 1] \mid \mathcal{H}_{t'} = h \Big] \\
	& \quad -\Mean_{\rho}\big[\delta \mid \mathcal{H}_{t'} = h \big] \times \I[\pi_{2,t}(h) = 1]  \bigg\} \times N_{1:T} \times (H + T - t')\\
	& - \Mean_{\rho} \Big[ \Mean_{\rho}\big[\delta \mid \mathcal{H}_{t'+1} \big]  \times \I[\pi_{2,t}(\mathcal{H}_{t'+1}) = 1] \mid \mathcal{H}_{t'} = h  \Big] \\
    & + \Mean_{\rho}\big[\delta \mid \mathcal{H}_{t'} = h \big] \times N_{1:t'+1, Tr} - c.
\end{align*}
\end{lemma}
Proof can be found in Appredix~\ref{appendix:infromation-gain}. 
The first term represents the value of information. 
Specifically, we have 
\begin{align*}
&\Mean_{\rho} \Big[  \Mean_{\rho}\big[\delta \mid \mathcal{H}_{t'+1} \big] \times \I[\pi_{2,t}(\mathcal{H}_{t'+1}) = 1] \mid \mathcal{H}_{t'} = h \Big]\\
 &\quad- \Mean_{\rho}\big[\delta \mid \mathcal{H}_{t'} = h \big] \times \I[\pi_{2,t}(h) = 1]\\
&=     \Mean_{\rho} \Big[  \Mean_{\rho}\big[\delta \mid \mathcal{H}_{t'+1} \big] \times \I \big[\Mean_{\rho}\big[\delta \mid \mathcal{H}_{t'+1} \big] > 0 \big] \mid \mathcal{H}_{t'} = h \Big]\\
&\quad - \Mean_{\rho}\big[\delta \mid \mathcal{H}_{t'} = h \big] \times \I \big[ \Mean_{\rho}\big[\delta \mid \mathcal{H}_{t'} = h \big] > 0 \big]\\
&\ge 0, 
\end{align*}
where the equality follows from the policy definitions. 
The inequality holds as long as $\Mean_{\rho}\Big\{ \Mean_{\rho}\big[\delta \mid \mathcal{H}_{t'+1} \big] \mid S_{t'}\Big\} = 
\Mean_{\rho}\big[\delta \mid S_{t'} \big]$, which is true since $\mathcal{H}_{t'+1}$ contains more information than $\mathcal{H}_{t'}$ does.

%% file: 3_Learning.tex
\subsection{Solving the Optimal Policy}\label{sec:optimization}
With the formulation above, the next step is to solve \eqref{eqn:objective} to obtain the optimal policy, so that we can use it for optimal termination. 
In the RL literature, there exists many efficient algorithms that be can be applied. 
For completeness,  in Section \ref{sec:single_MDP_solution}, we recap a classic one that we are working with. 
In Section \ref{sec:CDRL}, we further illustrate how to scale the solution so as to support real applications on major experimentation platforms.

\subsubsection{Policy Learning with RL}\label{sec:single_MDP_solution}

With an environment model, one can apply various state-of-the-art RL algorithms to solve the problem, such as the value-based algorithms(e.g., DQN \citep{mnih2015human}, Double DQN \citep{van2016deep}), 
policy-based algorithms (e.g., TRPO \citep{schulman2015trust}, PPO \citep{schulman2017proximal}) which directly learn a policy function, 
or actor-critic algorithms (e.g., A2C \citep{mnih2016asynchronous}, SAC \citep{haarnoja2018soft}, A3C \citep{mnih2016asynchronous}) which reduce the variance of policy-based methods by additionally learning the value function.  
Typically, the history can be summarized into a finite-dimensional vector. 
When not feasible, one can consider using models such as the long short-term memory network \citep{bakker2001reinforcement}.




With a discrete action set in our problem, we are currently using a classic value-based approach, deep Q-network \citep[DQN, ][]{mnih2015human}. 
For completeness, we briefly review its main idea here. 
DQN  uses a deep neural network to parameterize the value function $Q$, and it is a Q-learning-type algorithm based on the fixed-point iteration principle. 
It is motivated by the fact that $Q^{\pi^*} = \{Q_t^{\pi^*}\}_{t=1}^T$  is the unique solution of the \textit{Bellman optimality equation}
\begin{equation*}\label{eqn:bellman_Q}
    Q_t(a, h) = \Mean \Big(R_t + \gamma \argmax_{a \in \mathcal{A}_{t+1}} Q_{t+1}(a, \mathcal{H}_{t+1})  | A_t = a, \mathcal{H}_t = h \Big), \forall t. 
\end{equation*}
Regard the right-hand side as an operator on $Q = \{Q_t\}_{t=1}^T$. 
We can prove it is a contraction mapping and $Q^*$ is its fixed point, based on which we can derive an iterative algorithm. 
Details of the MDP-version of this  algorithm can be found in, e.g.,  Algorithm 1 in \cite{sutton2018reinforcement}. 
For completeness, we also present it in Appendix \ref{sec:num_details}.

\begin{remark}[Choice of the policy/value function class]
In our applications, we use neural networks as our value function class, due to their great representation power to approximate the true optimal policy (which is typically highly non-parametric and does not belong to a pre-defined function class). 
In use cases where the interpretability is desired, one can instead use, e.g., linear models. 
\end{remark}




\subsubsection{Contextual RL: Scalability and Deployment Feasibility}\label{sec:CDRL}
So far, we have described how to solve the optimal policy for one specific experiment. 
However, nowadays, an online experimentation platform in large companies may need to support tens of thousands of experiments per year. 
For different experiments, the NMDP models might be different (e.g., with different priors, different sample sizes,  etc.) and hence the optimal policies differ. 
Standard RL algorithms are designed for a fixed MDP environment, and its limited generalizability is well known as a huge bottleneck \citep{sodhani2021multi}. 

Therefore, one practical issue is the scalability and tractability of maintaining such an RL service. 
If we need to re-run the RL algorithm for every experiment, then such a service in not practical, as it requires significant manual effort. 
Even if we can trigger the training automatically, the computational resource needed and the training instability are of concerns. 


To address the scalability issue, we form the task as a contextual RL problem \citep{hallak2015contextual, benjamins2021carl}, a direction that is attracting increasing attention. 
Suppose each NMDP $\mathcal{M}_i = \{\mathcal{Y}^i_t, \mathcal{A}^i_t, \mathcal{P}^i_t, \mathcal{R}^i_t\}_{t=1}^T$  is associated with a feature vector $\vx_i$  (i.e., the so-called \textit{context}) and the difference between all NMDPs can be fully captured by the context, i.e., we can rewrite as $\mathcal{M}_i = 
\{\mathcal{Y}^{\vx_i}_t, \mathcal{A}^{\vx_i}_t, \mathcal{P}^{\vx_i}_t, \mathcal{R}^{\vx_i}_t\}_{t=1}^T$. 
Then, as long as we include the context as part of the history to define an \textit{augmented} history, we can construct one single new NMDP that unifies all these NMDPs (and generalize to those still unseen). 
For example, we can define the transition function $\mathcal{P}_t$ as $\mathcal{P}_t(h_t, \vx_i) = \mathcal{P}_t^{\vx_i}(h_t)$. 
Similar arguments apply to the other components. 



After this transformation, all the policy learning algorithms discussed in Section \ref{sec:single_MDP_solution} can be applied in the same way, except that we are going to sample trajectories by interacting with a set of MDPs in learn a generalizable policy. 
We can run training for only once to learn the policy, save it, and then apply it to any new experiments. 
Intuitively, this contextual policy learns how to act given both the history of an experiment and its features (the context). 
In Section \ref{sec:N-N}, we present a concrete example of the context vector.

%% file: 4_Numerical.tex
\section{Case Study on Amazon historical experiments}
\label{sec:experiment}
So far, we lay down a concrete but general framework for utility-maximizing early termination. 
In this section, we present an example of how we apply it to one of Amazon's largest experimentation platforms. 
We first introduce a specific transition model in Section \ref{sec:N-N}, and then present the meta-analysis results in Section  \ref{sec:meta-analysis}. 


\subsection{SDM Model with Independent Outcomes}\label{sec:N-N}


The SDM formulation in Section  \ref{sec:formulation} and the algorithms in Section \ref{sec:optimization} are general and not restricted to a specific outcome model. 
In this section, we provide one concrete form when the outcomes are independent across weeks. 
This model is what we are working with and the numerical results are also based on this model. 
Moreover, we can design appropriate state vector to make our NMDP model an MDP. 

As discussed in Section \ref{sec:state}, we will focus on discussing $p(Y_{t+1} \mid A_t = 0, \mathcal{H}_t, \vthe)$. 
We assume the observations in different weeks are independent. 
More formally, we assume that, conditioned on $\vthe$ and $\{A_t\}_{t=1}^T$, $\{Y_t\}_{t=1}^T$ are mutually independent. 
For simplicity of notations, we also assume the outcomes follow Gaussian distributions. 
However, we emphasize that, 
by the central limit theorem~\cite{casella2021statistical}, all derivations below still approximately hold with non-Gaussian outcomes, when the number of customers is large (typically the case). 
We assume a conjugate prior for $\mu_{g}$, the mean of group $g$. Therefore, our full model is 
\begin{eqnarray}
\label{eqn:conjugate}
	\mu_{g} &\sim& N(\mu_{0g}, \sigma_{0g}^2), \nonumber\\
	W_{i, t}|\mu_{g}  &\sim& N( \mu_{g}, \sigma_{g}^2), \textit{ for customer i in group} \; g \in \{\text{Tr, C}\}. 
\end{eqnarray}
Recall that $W_{i, t}$ are responses observed in week $t$ for customer $i$. 
In practice, the prior parameters $\mu_{0g}$ and $\sigma_{0g}$ are typically 
estimated via empirical Bayes \cite{maritz2018empirical} from historical experiments. 
Based on the derivations in Appendix~\ref{appendix:normal-normal}, following the Bayesian rule, for each group $g \in \{\text{Tr, C}\}$,  we have 
\begin{align*}
&\mu_{g}|  \bar{W}_{l, g}  
\sim N\bigg(\bigg( \frac{1}{\sigma_{0g}^2} + \frac{a_{g}(l)^2}{\sigma_{g}^2 b_{g}(l)} \bigg)^{-1}    \bigg( \frac{\mu_{0g}}{\sigma_{0g}^2} + \frac{a_{g}(l) \bar{W}_{l, g}}{\sigma_{g}^2 b_{g}(l)}\bigg)    ,\bigg( \frac{1}{\sigma_{0g}^2} + \frac{a_{g}(l)^2}{\sigma_{g}^2 b_{g}(l)} \bigg)^{-1}\bigg), \\
&\bar{W}_{l+1, g} | \mu_{g}, \bar{W}_{l, g} \sim N \bigg(
\frac{\bar{W}_{l, g}  N_{1:l, g} + \mu_{g}  N_{1:l+1, g} }{N_{1:l+1, g}}, \frac{1}{N_{1:l+1, g}} \sigma^2_{g}
\bigg). 
\end{align*}
Here, 
$\bar{W}_{l, g}$ is the average outcomes of $W_{i, 1:l}$ among customers who participated in the experiment up to week $l$, 
$a_g(l) = {c_g(l)} / {N_{1:l, g}}$,  
$b_g(l) =  {c_g(l)} / {N_{1:l, g}^2} $, 
and $c_g(l) =  \sum_{t = 1}^{l} N_{t, g}  \times (l- t + 1)$.



\textbf{MDP formulation.}
With this model, we can design appropriate state vector such that our model is an MDP. 
Note that, besides the posteriors at time $l$, the posterior at time $l+1$ only depends on the sample size, the data variance, and the priors. 
All of these components are regarded as known in our approach. 
Therefore, the history can be fully represented by the parameters of the posteriors at time $l$, i.e., $(\bar{W}_{l, T}, \bar{W}_{l, C})$. 
Hence, all information in the history can be summarized in the state vector $S_t = (S^b_t, I_t)$, where $S_l^b = (\bar{W}_{l, Tr}, \bar{W}_{l, C})^T$ and recall  that $I_t$ is the terminal indicator (see Section \ref{sec:action}). 
With the state vector defined, the full definition of the MDP then follows. 
The vector $(\mu_{0Tr}, \mu_{0C}, \sigma_{0Tr}, \sigma_{0C}, \sigma_{Tr}, \sigma_{C}, \{N_{t, Tr}\} ,\ \{N_{t, C}\} )$  contains all parameters needed to specify the environment and is the \textit{context vector} that we used in contextual RL (see Section \ref{sec:CDRL}).

\subsection{Meta-analysis}\label{sec:meta-analysis}


\textbf{Dataset and setup. } 
We collect all historical experiments that run on Amazon's largest experimentation platform in the past two years and have a length of at least $4$ weeks. 
Since we do not know the ground truth of the treatment effect (and hence the correct decision and the impacts), we cannot directly run a real data analysis~\footnote{In Appendix \ref{sec:real_heuristic}, we run on real data with a heuristic way of defining the ground truths. However, the set of results is only used for reference and should not be used to directly compare methods. }. 
Therefore, We design a real data-calibrated simulation study, 
where we calibrate a distribution of problem instances and study the average performance over them. 
We also present a simulation study in Appendix \ref{sec:pure_simu} to facilitate reproducibility, since we cannot share the real dataset and its more details due to confidentiality. 


For each historical experiment, we keep most variables (sample sizes, sample variances, etc.) the same as in the real trajectory. 
Regarding the opportunity cost, we first estimate the total saving if we can reduce one day for every experiment (details omitted due to business confidentiality), and then decompose these savings to each experiment based on their sample sizes.
We estimate the prior distributions using the empirical Bayes method from historical experiments. 
We set the maximum duration as $T=4$ weeks for all experiments. 
We take the first week of observations from the real  data, use the corresponding posterior to generate the   ground truth of the treatment effect, and then  simulate data based on the formulae in Section  \ref{sec:N-N}.

\input{tab_main}


\textbf{Baseline methods. } 
We compare the proposed RL framework with classic statistical methods for early termination. 
As reviewed in Section \ref{sec:related_work}, all these methods aim to control either the type-I error rate or FDR. 
For the alpha-spending approach (Frequentist)~\cite{demets1994interim, gordon1983discrete}, we use the O'Brien-Fleming spending function in the R package \textit{ldbounds}~\cite{ldbounds}  to control the type I error rate under $0.05$. 
For the sequential bayesian testing procedures ~\cite{jeffreys1961theory, schonbrodt2017sequential,deng2016continuous}, 
we consider three variants: 
(i) we compute the exact Bayes factor for one-sided hypothesis testing (our analysis shows that the one-sided test outperforms the two-sided one on this dataset), which we refer to as \name{BF}; 
(ii) we use the Posterior Odds (\name{POS})~\cite{deng2016continuous} in place of the Bayes factors, which also takes prior odds into consideration; 
(iii) we use the Jeffrey-Zellner-Siow (\name{JZS}) Bayes Factor implemented in the Python package \textit{Pingouin}~\cite{vallat2018pingouin}.
Following the guidelines ~\cite{jeffreys1961theory, schonbrodt2017sequential}, we try three threshold values ($3$, $10$ and $30$) for the three variants above. 
Lastly, we compare with the always valid p-values (\name{AVP}) approach \cite{johari2015always, johari2017peeking} based on a mSPRT test with type-I error constraint $0.05$. 
For the proposed RL method, we use a two-layer neural network with $128$ hidden nodes as the function class, and use the DQN implementation in \textit{Rllib}\citep{liang2018rllib}, an open-source RL package. 
More details can be found in Appendix~\ref{appendix:baseline}. 
We also compare with three fixed-horizon procedures, including the frequentist fixed horizon testing with level $0.05$ (\name{FFHT}), the 
Bayesian fixed horizon testing (\name{BFHT}) which rejects the null when the posterior probability of having a positive effect is larger than $0.66$, and the Bayesian fixed-horizon optimal decision rule (\name{BFHOD}) which recommends launch when the posterior mean of the gain is positive.




\textbf{Metrics.} We consider the following metrics for comparisons: 
\begin{enumerate}
		\item The percentage of experiments terminated early and the average number of weeks run for each experiment. 
   		\item False discovery rate (FDR; \#false positives / (\#false positives+\#true positives)), 
   		    power (proportion of correctly detecting significantly positive/negative effect when the true effect is indeed positive/negative), 
   		   and type-I error (false positive rate, i.e. the probability of mistakenly detecting significance when the true effect is not). 
    	\item The three components in Equation~\eqref{eqn:extended_utility_old}, including the opportunity cost, the impact during the experiment (relative to the control), and the launch impact (relative to the control). 
    	\item The average utility, which includes the three components above and is our main objective. 
    \end{enumerate}	



\textbf{Results. }
We present results from $50$ thousands trajectories in Table~\ref{tab:model_perform}. 
As expected, RL outperforms other methods and generates the highest average cumulative reward, i.e., RL maximizes customer experience. 
It is not surprising that most baseline methods have low type-I error rate or FDR. 
However, as mentioned in the introduction, most experiments in our applications are flat, therefore focusing on these metrics turns out to be too conservative. 
These limitations are particularly clear for frequentist methods or Bayesian early termination methods. 
The Bayesian fixed-horizon methods (\name{BFHT4} and \name{BFHOD4}), though have a great power and a high average reward, need too much opportunity costs since they cannot terminate experiments earlier. 
Finally, thanks to the contextual RL technique, the computational cost of making a decision for a new experiment is negligible (less than $0.005$ second). 
In conclusion, through the meta-analysis, we found that the proposed  approach achieves a desired balance between various metrics. 




\textbf{Policy behavior. } 
Recall that the learned policy is a neural network, mapping from the vector of (experiment's observations, experiment's features, time index) to the recommended optimal action. 
It is practically useful to provide interpretability of this policy to experimenters. 
One way to do that is by looking into how the recommended actions change with different inputs. 
To illustrate, we fix down one historical experiment at time $t=2$, then vary its opportunity cost and the posterior mean of ATE around the corresponding observed values while keeping the other variables fixed, and see how does the recommended action change. 



\begin{figure}[h]
\centering
     \includegraphics[width=0.35\textwidth]{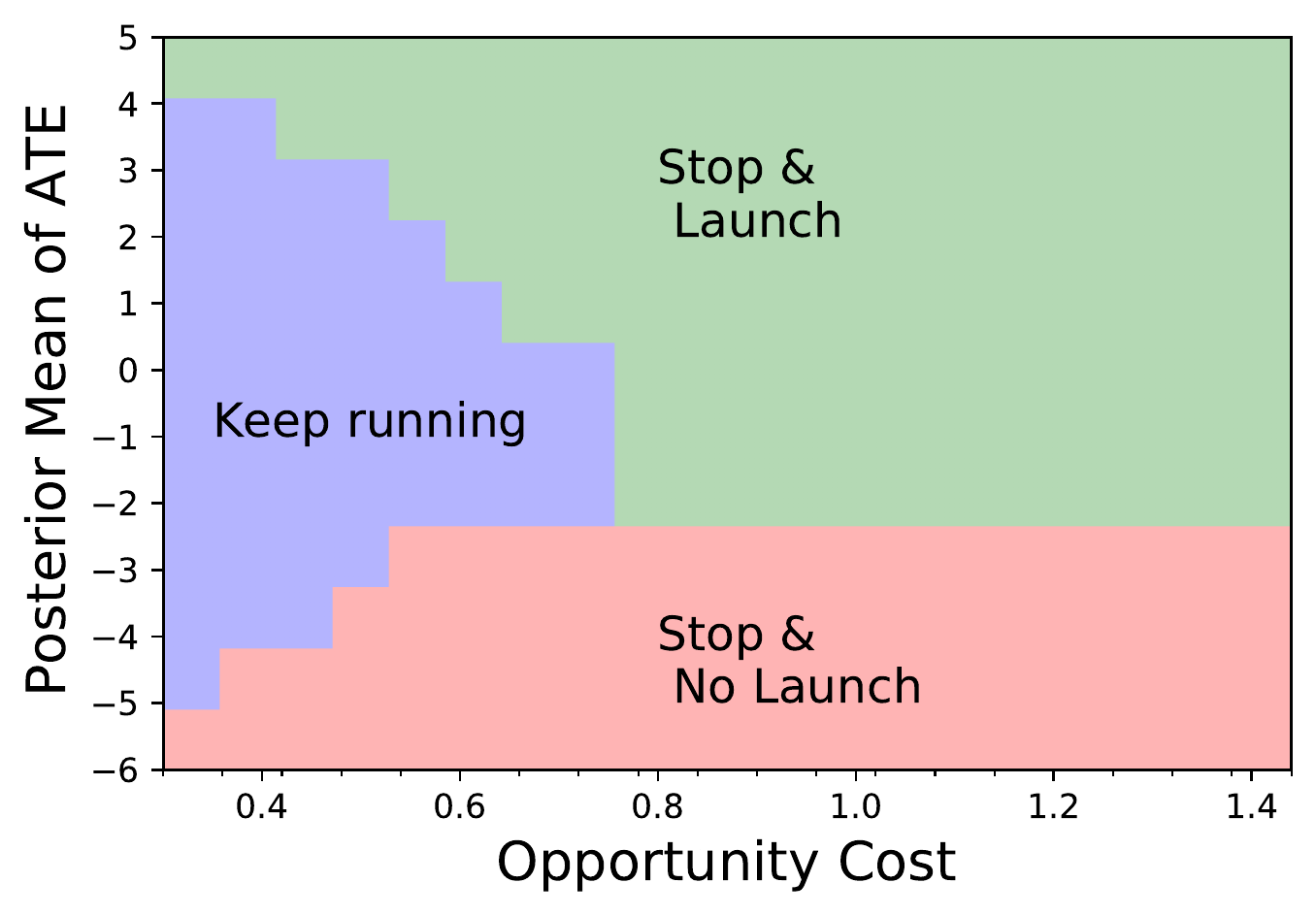} 
    \vspace{-.2cm}
\caption{Trend of the recommended action with the opportunity cost and the posterior mean of the treatment effect.
}
\label{fig:trend}
\end{figure}

We present the results in Figure \ref{fig:trend}, where the findings are overall reasonable. 
Along the y-axis, when the absolute value of the ATE posterior mean is away from zero, it implies the launch decision is less uncertain, and hence we observe the policy is more inclined towards “stop”. 
Along the x-axis, when the opportunity cost decreases, it implies the cost of keeping running the experiment decreases, hence we observe the policy is more inclined towards “keep running”. 
Similar findings on the relationship with the noise level of the data is presented in Appendix~\ref{appendix:noisy-mean}. 

%% file: tab_main.tex
\begin{table*}[t]
\caption{Meta-analysis results. 
The number after each method name indicates the tuning parameter being used. 
We omitted results with some tuning parameters that have very poor performance. 
Recall that all utility-related metrics share the same unit $D$, the meaning of which is omitted due to confidentiality. 
}
\label{tab:model_perform}
	\centering
\vspace{-.2cm}
	\renewcommand{\arraystretch}{1.5}
	\begin{tabular}{llllllllll}
			\hline
			Method
			 &  \makecell{\% Early \\ Terminated \\ Experiments}
			 &  \makecell{ Type I}
			 &  \makecell{ Power}  
			 &  \makecell{ FDR}
			 &  \makecell{Average \\ Weeks}
			 &  \makecell{Average \\  Opportunity\\ Cost (D)}
			 &  \makecell{ Average \\ Launch \\
				Impact (D)}
			 &  \makecell{Average \\ Experiment \\ Impact (D)} 
			 &  \makecell{ \textbf{Average} \\\textbf{Cumulative} \\\textbf{Reward} 
				 (D)
				 }\\
\hline
FFHT & 0.0\% & 0.0\% & 0.43 & 0.0\% & 4.0 & 2.63 & 4.31 & -0.0 & 1.68(0.07)\\
\hline
alpha-spending & 27.32\% & 0.0\% & 0.42 & 0.0\% & 3.64 & 2.48 & 4.24 & -0.0 & 1.83(0.07)\\
\hline
BFHT & 0.0\% & 0.03\% & 0.86 & 0.04\% & 4.0 & 2.63 & 6.28 & -0.0 & 3.64(0.07)\\
\hline
BFHOD  &  0.0\%  &  5.2\% & 0.95 & 5.4\% & 4.0 & 2.63 & 6.35 & -0.002 & 3.72(0.06)\\
\hline
BF 3 & 89.91\% & 0.39\% & 0.52 & 0.74\% & 1.7 & 0.69 & 4.4 & 0.04 & 3.91(0.06)\\
\hline
BF 10 & 61.65\% & 0.02\% & 0.47 & 0.05\% & 2.69 & 1.85 & 4.52 & 0.05 & 2.86(0.07)\\
\hline
POS 3 & 89.94\% & 0.39\% & 0.52 & 0.74\% & 1.7 & 0.68 & 4.39 & 0.04 & 3.9(0.06)\\
\hline
JZS 3 & 11.17\% & 0.0\% & 0.17 & 0.0\% & 3.84 & 2.61 & 1.67 & -0.0 & -0.91(0.05)\\
\hline
AVP & 16.15\% & 0.01\% & 0.41 & 0.02\% & 3.73 & 2.5 & 4.09 & -0.05 & 1.64(0.07)\\
\hline
\textbf{RL}  & 88.48\% & 7.01\% & 0.95 & 5.24\% & 2.31 & 0.53 &  5.18  &  -0.01  &  \textbf{4.64}(0.05)\\
\hline
\end{tabular}
\end{table*}

%% file: 5_Conclusion.tex
\vspace{-.2cm}
\section{Conclusion}\label{sec:conclusion}
In this paper, we propose an RL-based approach for continuously monitoring  online A/B experiments. 
Unlike traditional statistical approaches, 
our method aims to maximize the expected utility that consider a few different factors. 
We introduce in detail how we formulate this problem at Amazon, and also discuss how to solve the policy by RL algorithms. 
With a large-scale meta-analysis using past experiments from a large experimentation platform in Amazon, 
we find that the proposed approach leads to a significant gain in the expected utility. 


The task of making sequential decisions for online A/B experiments is challenging, as different experiments can vary a lot in their properties and it is not practical to tailor for tens of thousands of experiments. 
There are a few meaningful extensions of our framework. 
First, in some cases, the treatment effect is not homogeneous over time. 
We can model the uncertainty with a Bayesian time series model, from which we can solve a conservative and robust early termination rule. 
Second, it is practically meaningful to integrate statistical inference and optimal decision making. 
We propose a few approaches in Section \ref{sec:discussion_model}, which we will investigate numerically as our next step. 
Last, we estimate the priors from historical experiments, which can guarantee the optimality on average. 
It is useful to study the impact of the prior specifications.

%% file: 6_Appendix.tex
\appendix
\newpage

\section{Derivations}
\subsection{Predictive distribution for Normal-Normal model}
\label{appendix:normal-normal}
 By definition, the number of new customers enrolled in the experiment during week k and assigned to group $ * \in \{Tr, C\}*$ is represented by $N_{k, *}$. The observed activities of customer $i$ who first enrolled at week $k$ are represented by $W_{i, k}, .., W_{i, T}$. Sample mean of aggregated responses among customer participated in the experiments up to week $l$ is defined as:
	\begin{eqnarray*}
		\bar{W}_{l, *} &=& \frac{\sum_{i \in  \mathcal{I}_{1:l, *}} W_{i, 1:l}+ \sum_{i \in  \mathcal{I}_{2:l, *}} W_{i, 2:l}  ... + \sum_{i \in  \mathcal{I}_{l:l, *}} W_{i, l:l} }{\sum_{t = 1}^l N_{t, *}}\\
		&=&\frac{\sum_{t = 1}^l \sum_{i \in \mathcal{I}_{t:l, *} } W_{i, t:l, *} }{\sum_{t = 1}^l N_{t, *}}
	\end{eqnarray*}
	
	Following equation~\eqref{eqn:conjugate}, we had:
	\begin{eqnarray*}
		\bar{W}_{l, *} | \mu_* &\sim& N(\mu_* *a_*(l), \sigma_*^2 b_*(l))
	\end{eqnarray*}
	where $a_*(l) = \frac{c_*(l)}{N_{1:l, *}}$,  
$b_*(l) =  \frac{c_*(l)}{N_{1:l, *}^2} $, 
$c_*(l) =  \sum_{t = 1}^{l} N_{t, *}  \times (l- t + 1)$.

	Given $\bar{W}_{l, *}$ (one data point each arm at week $l$), posterior distributions of $\mu_*$ are:
	
	\begin{align*}
		&\mu_*|  \bar{W}_{l, *}  \propto exp\bigg(-\frac{1}{2} \bigg(\frac{\mu_* - \mu_{0*}}{\sigma_{0*}}\bigg)^2 \bigg)
		exp\bigg(-\frac{1}{2} \bigg(\frac{ \bar{W}_{l, *} - \mu_* *a_*(l)}{\sigma_* \sqrt{b_*(l)}}\bigg)^2 \bigg)\\
		&\sim N\bigg(\bigg( \frac{1}{\sigma_{0*}^2} + \frac{a_*(l)^2}{\sigma_*^2 b_*(l)} \bigg)^{-1}    \bigg( \frac{\mu_{0*}}{\sigma_{0*}^2} + \frac{a_*(l) \bar{W}_{l, *}}{\sigma_*^2 b_*(l)}\bigg)    ,\bigg( \frac{1}{\sigma_{0*}^2} + \frac{a_*(l)^2}{\sigma_*^2 b_*(l)} \bigg)^{-1}\bigg)
	\end{align*}

Model-based RL needs the ability to simulate the state transition, or roughly speaking, the capability to 
	simulate $\bar{W}_{l+1, *} | \bar{W}_{l, *} $.
	Note that 
	$f(\bar{W}_{l+1, *} | \bar{W}_{l, *}) 
	=\int f( \mu_* | \bar{W}_{l, *}) f(\bar{W}_{l+1, *} | \mu_*, \bar{W}_{l, *}) d \mu_*. 
	$
	Therefore, we only need to first sample $\mu_*$ from $\mu_* | \bar{W}_{l, *}$, and then sample 
	$\bar{W}_{l+1, *}$ from $\bar{W}_{l+1, *} | \mu_*, \bar{W}_{l, *}$. 
	We have 
	\begin{eqnarray*}
		\bar{W}_{l+1, *} | \mu_*, \bar{W}_{l, *} &\sim& N \bigg(
		\frac{\bar{W}_{l, *}  * N_{1:l, *} + \mu_T * N_{1:l+1, *} }{N_{1:l+1, *}}, \frac{1}{N_{1:l+1, *}} \sigma^2_{*}
		\bigg). 
	\end{eqnarray*}

\subsection{Role of information Gain}
\label{appendix:infromation-gain}
Proof the fixed-horizon Bayesian decision rule:
Consider a fixed-horizon Bayesian optimal decision scenario where a final decision is made at $T$, if is no opportunity cost for running the experiment (i.e., $c = 0$) and ignore the running experiment impact ( third term in Eq~\ref{eqn:objective}), then the objective becomes:
\begin{eqnarray*}
&&\Mean_{\rho}^\pi \Big[\sum_{t=1}^T R_t  | \mathcal{H}_1 = h \Big] = \Mean_{\rho}^\pi \Big[R_T | \mathcal{H}_1 = h \Big]\\
&=& \Mean_{\rho}\Big[ ( \Mean_{\rho}(\delta \mid \mathcal{H}_T) \times H \times N - c_h) \times \I\big[\pi_T(\mathcal{H}_T) = 1 \big] \mid \mathcal{H}_1 = h \Big]
\end{eqnarray*}
where the first equality is due to that only the final decision is allowed, and the second equality is from our reward definitions above. The optimal policy is hence $\pi_T(\mathcal{H}_T) = 2 - \I\big[  \Mean_{\rho}(\delta \mid \mathcal{H}_T ) \times H \times N> c_h \big]$

Proof of Lemma~\ref{lemma:reward-diff}
\begin{eqnarray*}
	&&\Mean_{\rho}^{\pi_2} \Big[\sum_{t=t'}^T R_t \mid \mathcal{H}_{t'} = h  \Big] - \Mean_{\rho}^{\pi_1} \Big[\sum_{t=t'}^T R_t \mid \mathcal{H}_{t'} = h \Big]\\
	&=& \Mean_{\rho}^{\pi_2} \Big[ R_t'  +R_{t'+1} \mid \mathcal{H}_{t'} = h  \Big] - \Mean_{\rho}^{\pi_1} \Big[ R_t' \mid \mathcal{H}_{t'} = h \Big] \\
	&=& \Mean_{\rho}^{\pi_2} \Big[ R_{t'+1} \mid \mathcal{H}_{t'} = h  \Big] - \Mean_{\rho}^{\pi_1} \Big[ R_t' \mid \mathcal{H}_{t'} = h \Big] \\
	&-& c + \Mean_{\rho}^{\pi_1} \Big[ R_t' \mid \mathcal{H}_{t'} = h \Big] *N_{1:t'+1, Tr} \\
	&& \text{(Plug in Equation~\eqref{eq:reward})}\\
	&= &
	\Mean_{\rho} \Big[ \big( \Mean_{\rho}\big[\delta \mid \mathcal{H}_{t'+1} \big] \times (H + T - t' - 1) \times N \big)\\
	&& \quad \times \I[\pi_{2,t}(\mathcal{H}_{t'+1}) = 1] \mid \mathcal{H}_{t'} = h \Big] \\
	 && \quad- \Mean_{\rho}\big[\delta \mid \mathcal{H}_{t'} = h \big] \times (H + T - t' ) \times N \times \I[\pi_{1,t}(h) = 1] \\
	&& \quad - c  + \Mean_{\rho}^{\pi_1} \Big[ R_t' \mid \mathcal{H}_{t'} = h \Big] *N_{1:t'+1, Tr}\\
	&=&
	\bigg\{\Mean_{\rho} \Big[  \Mean_{\rho}\big[\delta \mid \mathcal{H}_{t'+1} \big] \times \I[\pi_{2,t}(\mathcal{H}_{t'+1}) = 1] \mid \mathcal{H}_{t'} = h \Big] \\
	&&\quad - \Mean_{\rho}\big[\delta \mid \mathcal{H}_{t'} = h \big] \times \I[\pi_{2,t}(h) = 1]  \bigg\} \times N \times (H + T - t')\\
	&& \quad - \Mean_{\rho} \Big[ \Mean_{\rho}\big[\delta \mid \mathcal{H}_{t'+1} \big]  \times \I[\pi_{2,t}(\mathcal{H}_{t'+1}) = 1] \mid \mathcal{H}_{t'} = h  \Big]  \\
	&& \quad - c + \Mean_{\rho}^{\pi_1} \Big[ R_t' \mid \mathcal{H}_{t'} = h \Big] *N_{1:t'+1, Tr}.
\end{eqnarray*}



\section{Additional numerical details and results}
\subsection{Pseudo-code of Q-learning}\label{sec:num_details}
In Algorithm \ref{alg:DQN}, we present the pseudo-code of the Q-learning algorithm that we used for our experiment. 
\input{DQN}

\subsection{Trend wth noise levels}
\label{appendix:noisy-mean}

\begin{figure}[h]
\centering
     \includegraphics[width=0.4\textwidth]{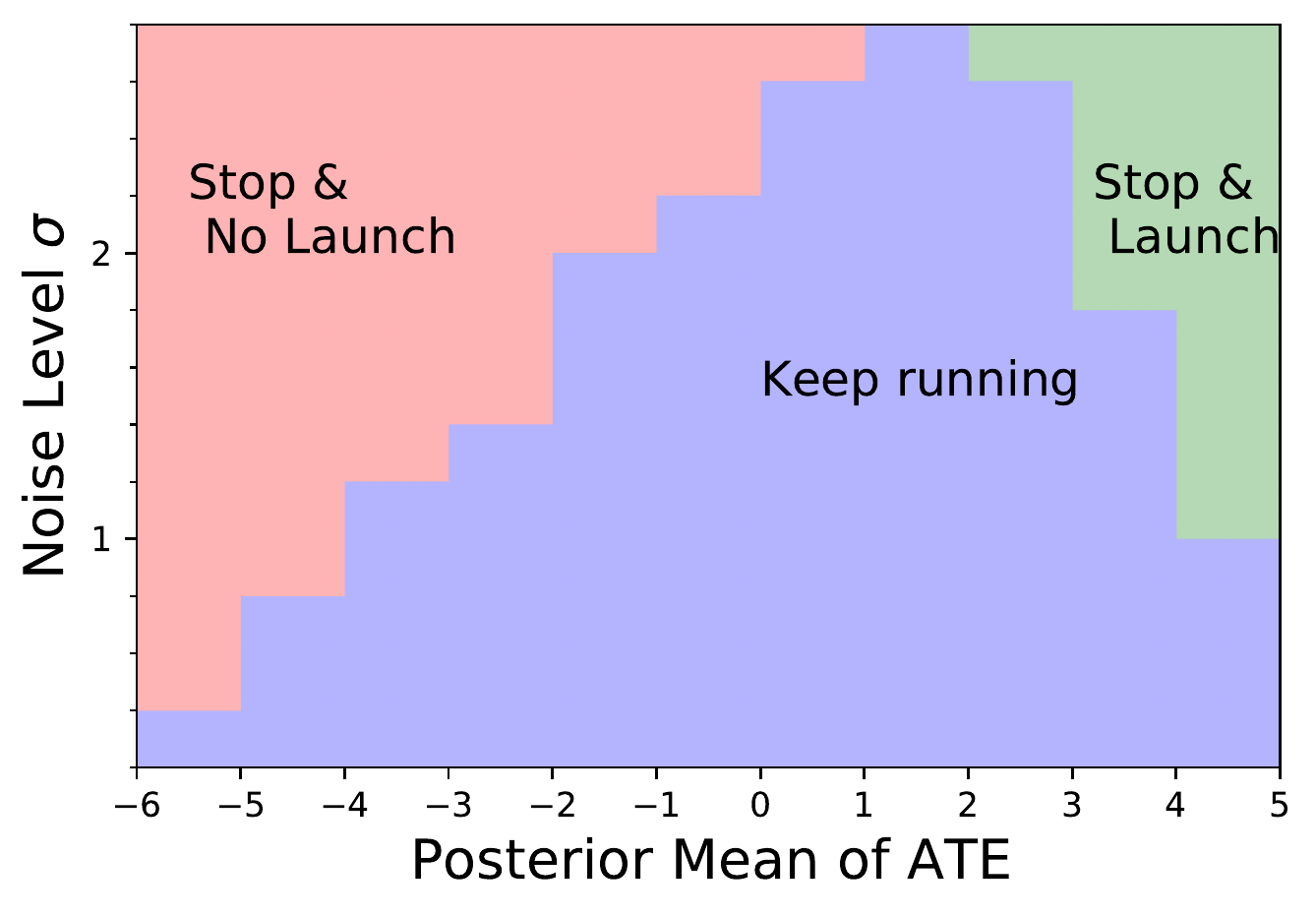} 
\caption{Trend of the recommended action with the noise level of the data and the posterior mean of the estimated treatment effect. 
}
\label{fig:trend2}
\end{figure}

Figure~\ref{fig:trend2} shows that when the noise of future data points decreases (i.e., information value increases), the optimal policy generated from RL framework will be more inclined to "keep running". 
Similarly, we also observed that when the current posterior variance decreases, RL inclined to terminating earlier. 
These two trends together show that our framework can utilize the progress from other research topics that reduce the uncertainty (e.g., covariate adjustment). 

\subsection{Real data analysis with heuristic ground truths}\label{sec:real_heuristic}

\input{tab_real}

In this section, we directly run different methods on historical trajectories. 
However, Since we do not know the ground truth of the treatment effect (and hence the correct decision and the impacts), it is impossible to evaluate various methods. 
We adopt a heuristic approach that uses the posterior mean after $4$ weeks as the ground truth. 
Accordingly, \textit{all decision accuracy-related metrics (type-I error, power, FDR) should be understood as the differences with the fixed-horizon Bayesian decision rule, and all utility-related metrics are measured using the posterior mean. }
We emphasize that, this is a heuristic approach to provide users a sense of what may happen compared with running the full horizon of experiments, and we do not recommend over-interpreting this set of results. 

We apply an affine transformation on the utility-related components and use another unit $E$, due to confidentiality consideration. 
This does not affect the conclusions. 
We do not compare with fixed-horizon Bayesian procedure, as by design they are directly based on the ground truths. 
From the results, we can see that RL yields significantly higher (empirical) utility, consistent with our design. 
In particular, as mentioned in the introduction, the signal in our applications is typically weak and existing methods are too conservative.

\subsection{Simulated data analysis}\label{sec:pure_simu}
\input{tab_pure_simu}
We simulate 3000 experiment trajectories and compare the performance of proposed RL model with baseline methods in Table~\ref{tab:model_perform_simulate}. 
The data generation process is as follows. 
\begin{itemize}
\item Sample size. We assume the number of customers shown in each week follows the beta-geometric model. (~\cite{richardson2022bayesian} without the censoring in the likelihood function.) The parameters $\alpha$ and $\beta$ of the beta distribution were randomly simulated from a uniform distribution, with $\alpha \sim Uniform(0.1, 1)$ and  $\beta \sim Uniform(4, 60)$. 
Assume the total number of customers is 10K.
\item Sample weekly responses for 4 weeks. We followed the normal-normal model in ~\eqref{eqn:conjugate}, with parameters as 
$\mu_{0C}$ = 0.1, $\sigma_{0C}$ = 2, 
$\mu_{0Tr}$ = 0.1, $\sigma_{0Tr}$ = 2.83 and $\sigma_C = \sigma_T = 100$. 
\item The total weekly opportunity cost is $1.5 \times 10^8$, decomposed to each experiment according to their sample sizes. Huddle costs are set as $0$. 

\end{itemize}

\section{Baseline Methods}
\label{appendix:baseline}
In this section, we provide the details  of the baseline models we use and how we implement them. 
For simplicity, the notations used in this section slightly differs with those in the main paper. 

We first give a more formal description of the continuous monitoring problem. 
A continuous monitoring procedure differs from fixed horizon testing by using a set of function $\tau_1(\mathcal{H}_1), \tau_2(\mathcal{H}_2) \ldots$, each of which is a function mapped to $\{stop, continue\}$. 
Define $\tau$ as the stopping time, i.e.,  the first time when $\tau_t(\mathcal{H}_t)$ is "stop".
The continuous monitoring procedure also contains 
a list of decision rules $\bm{\eta} = \{ \eta_1(\mathcal{H}_1), \eta_2(\mathcal{H}_2) ... \}$, 
and it makes the final launch decision at time $\tau$ as $ \eta_{\tau}(\mathcal{H}_1)$. 

\subsection{Alpha spending function approach}
		\label{appendix:alpha}
	An important work is the alpha spending function approach~\cite{demets1994interim, gordon1983discrete}.  In this approach, the cumulative Type I error rate is "spent" across the interim tests according to a pre-specified function $\alpha(t)$.
	If $T$ tests are planned to be conducted weekly, at the end of each week, $i_k$ represents the information available at the $k^{th}$ interim test, $k = 1, 2, ... T$. The information fraction is $t_k^* = i_k / I$, where $I$ is the total information. The decision boundary can be determined successively by following $T$ equations:
	\begin{eqnarray*}
		P(|Z(1)| \geq Z_c(1), or |Z(2)| \geq Z_c(2), \ldots, or |Z(k)| \geq Z_c(k)) = \alpha (t^*_k)
	\end{eqnarray*}
	where $Z(1), Z(2), ...Z(k) $ are test statistics from the k interim analyses and $Z_c(1), .., Z_c(k)$ are the boundary values. 
	We use the approximated O'Brien-Fleming function (2-sided case) which generates a conservative boundary at the beginning:
	$ \alpha (t) = 4(1-\phi(\phi^{-1}(1 - \alpha/4)/\sqrt{t}))$.
	 where $\phi$ is the standard normal CDF.  To solve the boundary values, joint distribution of $Z(1), ..., Z(k)$ are required. 
	 Details can be referred to the FORTRAN program "ld98"~\cite{ld98} and R package "ldbounds"~\cite{ldbounds}. 

\subsection{Sequential bayes factor}\label{appendix:bayes_factor}
First, we reviewed the Bayesian hypothesis testing~\cite{deng2015objective}. Assume $\bar{Y}|\mu \sim N(\mu, \sigma^2)$ where $\bar{Y} = \bar{Y}_{Tr} - \bar{Y}_C$ (ignore subscript $t$ for the simplicity), $\sigma^2 = \sigma_C^2/n_C + \sigma_{Tr}^2/n_T$ and $\mu = \mu_T - \mu_C$ to compare the mean of two groups. 

We use the one-sided test. Null and alternative hypotheses are defined as:
\begin{eqnarray}
H_0: \mu \leq  0 &\quad& \mu \sim \mathcal{I}[\mu < 0 ]N(\mu_0, \sigma^2_0) \nonumber \\
H_1:  \mu > 0 &\quad&
\mu \sim \mathcal{I}[\mu > 0 ] N(\mu_1, \sigma^2_1) 
\label{eqn:hypothesis}
\end{eqnarray}

where $\mu = \mu_{Tr} - \mu_C$ and $\mathcal{I}[\mu > 0 ] N$ represents the truncated normal. The null and alternative have distinct prior distribution for $\mu$ and all variance terms $\sigma_*^2$ are assumed known.

Following the Bayes's Rule, the posterior odds is defined as:

\begin{equation}
	\text{PostOdds} := \frac{P(H_1 | \bm{Y})}{P(H_0 | \bm{Y})} = \frac{P(H_1)}{P(H_0)}\frac{ P(\bm{Y}| H_1)}{P(\bm{Y}|H_0)} 
	\label{eqn:posterior_odds}
\end{equation}
$ \frac{P(H_1)}{P(H_0)}$ is referred to as Prior Odds and $B_{10}:=\frac{ P(\bm{Y}|H_1)}{P(\bm{Y}| H_0)}$  is referred to as the Bayes Factor~\cite{schonbrodt2017sequential, kass1995bayes,dienes2011bayesian}.
Decisions can be based on Bayes factors or the posterior odds. Jeffrey~\cite{jeffreys1961theory, schonbrodt2017sequential} suggested the relationship between the scale of Bayes factors  and the strength of the evidence as follows: $1 < BF < 3 $, the evidence is considered anecdotal, $3 < BF < 10$ represents moderate evidence, $10 < BF < 30$ is strong evidence, and $BF > 30$ is very strong evidence.

From:
\begin{eqnarray*}
	&\int& p(
	\bar{y}| \mu)   p(\mu | H_0) d \mu   = 	\frac{1}{ \Phi(\frac{0 - \mu_0}{\sigma_0} ) - \Phi(\frac{-\infty - \mu_0}{\sigma_0} ) } 
	\frac{1}{ 2\pi \sigma  \sigma_0} \\
	&&\int_{-\infty}^0 exp\bigg( -\frac{1}{2} \bigg(  \frac{(\bar{y} - \mu)^2}{\sigma^2}  + \frac{( \mu - \mu_0)^2}{ \sigma_0^2} \bigg) \bigg) d\mu \\
	&=&\frac{1}{ \Phi(\frac{-\mu_0}{\sigma_0} )  } \frac{1}{ 2\pi \sigma  \sigma_0}   exp\bigg( -\frac{1}{2} \frac{(\mu_0 - \bar{y})^2}{ \sigma_0^2 + \sigma^2}  \bigg)\\
	&&\int_{-\infty}^0 exp \bigg( -\frac{1}{2} \frac{ \sigma_0^2 +\sigma^2}{ \sigma_0^2\sigma^2} \bigg(\mu - \frac{\bar{y}  \sigma^2_0 + \mu_0 \sigma^2}{ \sigma^2_0 + \sigma^2} \bigg)^2 \bigg) d\mu \\
	&=&\frac{1}{\sqrt{2\pi ( \sigma_0^2 +\sigma^2} )}exp\bigg( -\frac{1}{2} \frac{(\mu_0 - \bar{y})^2}{ \sigma^2 + \sigma_0^2}  \bigg) \frac{\Phi(- \mu_0'/\sigma_0')}{\Phi( -\mu_0/\sigma_0) } \\
\end{eqnarray*}
where $\mu' = \frac{\bar{y}  \sigma^2_0 + \mu_0 \sigma^2}{ \sigma^2_0 + \sigma^2} $, $\sigma'_0 = \sqrt{\frac{1}{\sigma^2 + \sigma_0^2}}$

The exact Bayes Factor for one-sided test in~\eqref{eqn:hypothesis} is:
\begin{eqnarray*}
	BF[H_1 : H_0] &=& \frac{P(\bar{y}| H_1)}{P(\bar{y}|H_0)}  \\
	&=&\frac{\int _{\mu > 0} p(\bar{y}| \mu) p(\mu | H_1) d \mu}{\int _{\mu < 0} p(\bar{y}| \mu) p(\mu | H_0) d \mu}\\	&=&\frac{1-\Phi(-\mu'_1/\sigma'_1)}{1-\Phi(-\mu_1/\sigma_1)}* \frac{\Phi(-\mu_0/\sigma_0)}{\Phi(-\mu'_0/\sigma'_0)}\sqrt{\frac{\sigma^2_0 + \sigma^2}{\sigma^2_1 + \sigma^2}}\\
	&&exp\bigg( -\frac{1}{2} \frac{(\mu_1 - \bar{y})^2}{ \sigma^2 + \sigma_1^2} +\frac{1}{2} \frac{(\mu_0 - \bar{y})^2}{ \sigma^2 + \sigma_0^2}   \bigg)
\end{eqnarray*}	

\subsection{Sequential bayes factor   with posterior odds}\label{sec:Pos_odds}
\cite{deng2016continuous} proves the validity of Bayesian testing with continuous monitoring following the proposed stopping rule. Given a stopping rule $\tau$ which decides whether to continue or stop on the basis of only the present and past events and $\tau$ is finite almost surely, Theorem1 \cite{deng2016continuous} proves that the interpretation of posterior odds remains unchanged with optional stopping at $\tau$. 
Thus, ~\cite{deng2016continuous} claims that it is natural to define the  stopping rule using the false discovery rate (FDR) bound as the following:
\begin{itemize}
	\item[a.] Define FDR bound $1/(K+1)$.
	\item[b.] Stop and reject $H_0$ when the posterior odds defined in Equation~\eqref{eqn:posterior_odds} is equal to or greater than $K$. In a symmetric design, early stopping can also be done to accept $H_0$ for futility if the posterior odds is equal to or less than $1/K$
\end{itemize}

\subsection{Always valid inference}
\label{appendix:AVP}
~\cite{johari2015always} proposed the idea of "always valid inference" and derived the always-valid p-value for mixture sequential ratio probability test (mSPRTs)~\cite{robbins1970statistical}. 
The mSPRT is parameterized by a mixing distribution $H$ over $\Theta_1$. Given the observed sample average $\bar{y}_n $ up to time $n$, the mixture likelihood ratio with respect to $H$ is defined as:
\begin{equation}
	\label{eq:msprt}
	\Lambda_n^{H}(\bar{y}_n) = \int_{\Theta} \biggl(\frac{f_{\theta}(\bar{y}_n)}{f_{\theta_0}(\bar{y}_n)} \biggr)^n dH(\theta)
\end{equation}

$\Lambda_n^{H}(\bar{y}_n)$ intuitively, is the evidence against $H_0$ in favor of alternative $H$ based on the first n observations.

Now apply it to two group mean-difference case and follow the one side hypothesis testing in Equation~\eqref{eqn:hypothesis}. The derivation for $\Lambda_n^{H}(\bar{y}_n)$  is similar to that in Appendix~\ref{appendix:bayes_factor}. 
Early stop if $\Lambda_n^{H}(\bar{y}_n)  \leq 1/\alpha$.

%% file: DQN.tex
\vspace{.2cm}
 \begin{algorithm}[!ht]
\KwData{
horizon $T$, 
number of episodes $M$
, exploration probability $\{\epsilon_e\}$
, the function class for the Q-function
}

\textbf{Initialization:}
set the replay buffer $\mathcal{D}$ as empty, and initialize the action-value function $\{Q_t(\cdot, \cdot)\}$ with random weights

\For{episode $e = 1, \dots, M$}{
    Randomly initialize with an appropriate state $S_0$, sample one time period of data, and compute $S_1$
    
    \For{$t = 1, \dots, T$}{
        With probability $\epsilon_e$ select a random action $A_t \in \mathcal{A}_t$; otherwise select $A_t = \argmax_{a \in \mathcal{A}_t} Q_t(S_t, a)$
        
        Take action $A_t$, 
        sample the next state $S_{t+1}$ according to our transition model, 
        and generate the reward $R_t$
        
        Store transition $(S_t, A_t, R_t, S_{t+1}, t)$ in $\mathcal{D}$
        
        Sample a mini-batch $\{(s_i, a_i, r_i, s'_i, t_i)\}$ from $\mathcal{D}$
        
        For each data point in the mini-batch, set $y_i = r_i + \max_{a \in \mathcal{A}_{t_i+1}} Q_{t_i + 1}(s'_i, a)$ 
        
        Perform a gradient descent step on $Q_t$ using the training dataset $\{(s_i, y_i) : t_i = t\}, \forall t$
    }
}

 \KwResult{
 optimal policy $\hat{\pi} = (\hat{\pi}_0, \dots, \hat{\pi}_T)$, where
 $\hat{\pi}_{t}(\cdot) = \argmax_{a \in \mathcal{A}_t} Q_t(\cdot, a)$. 
 }  
 \caption{
 Q-learning with experience replay and epsilon-greedy exploration for finite-horizon planning.}\label{alg:DQN}
\end{algorithm}
\vspace{.5cm}


%% file: tab_real.tex
\begin{table*}[t]
\caption{Meta-analysis results on real experiments with heuristic-based ground truths. 
The number after each method name indicates the tuning parameter being used. 
We omitted results with some tuning parameters that have very poor performance. 
}
\label{tab:model_perform}
	\centering
\vspace{-.2cm}
	\renewcommand{\arraystretch}{1.5}
	\begin{tabular}{llllllllll}
			\hline
			Method
			 &  \makecell{\% Early \\ Terminated \\ Experiments}
			 &  \makecell{ (Empirical) \\ Type I}
			 &  \makecell{ (Empirical) \\ Power}  
			 &  \makecell{ (Empirical) \\ FDR}
			 &  \makecell{Average \\ Weeks}
			 &  \makecell{(Empirical) \\Average \\  Opportunity\\ Cost (E)}
			 &  \makecell{ (Empirical) \\Average \\ Launch \\
				Impact (E)}
			 &  \makecell{(Empirical) \\ Average \\ Experiment \\ Impact (E)} 
			 &  \makecell{ (Empirical) \\ Average \\Cumulative \\ Reward
				 (E)
				 }\\
\hline
FFHT & 0.0\% & 0.0\% & 0.05 & 0.0\% & 4.0 & 0.31 & 0.58 & -0.02 & 0.25(0.12)\\
\hline
alpha-spending & 1.94\% & 0.0\% & 0.05 & 0.0\% & 3.97 & 0.31 & 0.58 & -0.02 & 0.25(0.12)\\
\hline
BF 3 & 83.68\% & 0.9\% & 0.05 & 15.0\% & 1.71 & 0.0 & 0.15 & 0.01 & 0.16(0.03)\\
\hline
BF 10 & 57.73\% & 0.15\% & 0.01 & 10.0\% & 2.51 & 0.02 & 0.07 & 0.02 & 0.07(0.02)\\
\hline
BF 30 & 32.32\% & 0.08\% & 0.0 & 14.29\% & 3.23 & 0.06 & 0.02 & 0.03 & -0.02(0.01)\\
\hline
POS 3 & 84.62\% & 0.83\% & 0.05 & 14.86\% & 1.68 & 0.0 & 0.15 & 0.01 & 0.16(0.03)\\
\hline
JZS 3 & 0.43\% & 0.0\% & 0.0 & 0.0\% & 3.99 & 0.31 & 0.01 & -0.02 & -0.32(0.02)\\
\hline
AVP & 0.31\% & 0.15\% & 0.01 & 22.22\% & 3.99 & 0.31 & 0.03 & -0.02 & -0.3(0.03)\\
\hline
\textbf{RL}  & 98.96\% & 27.48\% & 0.61 & 32.62\% & 1.81 & 0.03 &  1.38 &  -0.01  &  \textbf{1.34}(0.32)\\
\hline
\end{tabular}
\end{table*}

%% file: tab_pure_simu.tex
\begin{table*}[t]
\caption{Simulation results. 
}
\label{tab:model_perform_simulate}
	\centering
\vspace{-.2cm}
	\renewcommand{\arraystretch}{1.5}
	\begin{tabular}{llllllllll}
			\hline
			Method
			 &  \makecell{\% Early \\ Terminated \\ Experiments}
			 &  \makecell{  \\ Type I}
			 &  \makecell{  \\ Power}  
			 &  \makecell{  \\ FDR}
			 &  \makecell{Average \\ Weeks}
			 &  \makecell{ \\Average \\  Opportunity\\ Cost (E)}
			 &  \makecell{  \\Average \\ Launch \\
				Impact (E)}
			 &  \makecell{\\ Average \\ Experiment \\ Impact (E)} 
			 &  \makecell{ \\ Average \\Cumulative \\ Reward
				 (E)
				 }\\
\hline
FFHT 4 & 0.0\% & 0.0\% & 0.0 & 0.01\% & 4.0 & 0.15 & 0.24 & -0.0 & 0.1(0.01)\\
\hline
alpha-spending & 15.23\% & 0.01\% & 0.0 & 0.02\% & 3.82 & 0.14 & 0.24 & -0.0 & 0.11(0.01)\\
\hline
BFHT & 0.0\% & 0.1\% & 0.01 & 0.13\% & 4.0 & 0.15 & 0.39 & -0.0 & 0.24(0.01)\\
\hline
BF 3 & 90.83\% & 0.49\% & 0.01 & 0.38\% & 1.64 & 0.03 & 0.26 & -0.0 & 0.24(0.02)\\
\hline
BF 10 & 61.27\% & 0.24\% & 0.01 & 0.25\% & 2.6 & 0.08 & 0.35 & -0.0 & 0.28(0.02)\\
\hline
BF 30 & 38.3\% & 0.1\% & 0.01 & 0.15\% & 3.22 & 0.11 & 0.34 & -0.0 & 0.24(0.02)\\
\hline
POS 3 & 90.4\% & 0.48\% & 0.01 & 0.37\% & 1.66 & 0.03 & 0.26 & -0.0 & 0.24(0.02)\\
\hline
JZS 3 & 7.2\% & 0.0\% & 0.0 & 0.01\% & 3.89 & 0.14 & 0.13 & -0.0 & -0.01(0.01)\\
\hline
AVP & 33.37\% & 0.14\% & 0.01 & 0.18\% & 3.24 & 0.11 & 0.35 & -0.01 & 0.25(0.02)\\
\hline
\textbf{RL} &  98.90\% & 25.36\% & 0.72 & 27.37\% & 2.33 & 0.06 & 0.38 & -0.0 & 0.32(0.02)\\
\hline
\end{tabular}
\end{table*}


%% file: 00_KDD_main.bbl

\begin{thebibliography}{54}


\ifx \showCODEN    \undefined \def \showCODEN     #1{\unskip}     \fi
\ifx \showDOI      \undefined \def \showDOI       #1{#1}\fi
\ifx \showISBNx    \undefined \def \showISBNx     #1{\unskip}     \fi
\ifx \showISBNxiii \undefined \def \showISBNxiii  #1{\unskip}     \fi
\ifx \showISSN     \undefined \def \showISSN      #1{\unskip}     \fi
\ifx \showLCCN     \undefined \def \showLCCN      #1{\unskip}     \fi
\ifx \shownote     \undefined \def \shownote      #1{#1}          \fi
\ifx \showarticletitle \undefined \def \showarticletitle #1{#1}   \fi
\ifx \showURL      \undefined \def \showURL       {\relax}        \fi
\providecommand\bibfield[2]{#2}
\providecommand\bibinfo[2]{#2}
\providecommand\natexlab[1]{#1}
\providecommand\showeprint[2][]{arXiv:#2}

\bibitem[{ Charlie Casper, Thomas Cook and Oscar A. Perez.}(2 02)]%
        {ldbounds}
\bibfield{author}{\bibinfo{person}{{ Charlie Casper, Thomas Cook and Oscar A.
  Perez.}}} \bibinfo{year}{2022-02}\natexlab{}.
\newblock \bibinfo{booktitle}{\emph{An R Package for Group Sequential
  Boundaries Using Alpha Spending Functions}}.
\newblock
\newblock
\shownote{\url{https://cran.r-project.org/web/packages/ldbounds/index.html}}.


\bibitem[Anderson(2016)]%
        {anderson2016library}
\bibfield{author}{\bibinfo{person}{Linda Anderson}.}
  \bibinfo{year}{2016}\natexlab{}.
\newblock \showarticletitle{Library website visits and enrollment trends}.
\newblock \bibinfo{journal}{\emph{Evidence Based Library and Information
  Practice}} \bibinfo{volume}{11}, \bibinfo{number}{1} (\bibinfo{year}{2016}),
  \bibinfo{pages}{4--22}.
\newblock


\bibitem[Andrychowicz et~al\mbox{.}(2020)]%
        {andrychowicz2020learning}
\bibfield{author}{\bibinfo{person}{OpenAI:~Marcin Andrychowicz},
  \bibinfo{person}{Bowen Baker}, \bibinfo{person}{Maciek Chociej},
  \bibinfo{person}{Rafal Jozefowicz}, \bibinfo{person}{Bob McGrew},
  \bibinfo{person}{Jakub Pachocki}, \bibinfo{person}{Arthur Petron},
  \bibinfo{person}{Matthias Plappert}, \bibinfo{person}{Glenn Powell},
  \bibinfo{person}{Alex Ray}, {et~al\mbox{.}}} \bibinfo{year}{2020}\natexlab{}.
\newblock \showarticletitle{Learning dexterous in-hand manipulation}.
\newblock \bibinfo{journal}{\emph{The International Journal of Robotics
  Research}} \bibinfo{volume}{39}, \bibinfo{number}{1} (\bibinfo{year}{2020}),
  \bibinfo{pages}{3--20}.
\newblock


\bibitem[Bakker(2001)]%
        {bakker2001reinforcement}
\bibfield{author}{\bibinfo{person}{Bram Bakker}.}
  \bibinfo{year}{2001}\natexlab{}.
\newblock \showarticletitle{Reinforcement learning with long short-term
  memory}.
\newblock \bibinfo{journal}{\emph{Advances in neural information processing
  systems}}  \bibinfo{volume}{14} (\bibinfo{year}{2001}).
\newblock


\bibitem[Benjamins et~al\mbox{.}(2021)]%
        {benjamins2021carl}
\bibfield{author}{\bibinfo{person}{Carolin Benjamins}, \bibinfo{person}{Theresa
  Eimer}, \bibinfo{person}{Frederik Schubert}, \bibinfo{person}{Andr{\'e}
  Biedenkapp}, \bibinfo{person}{Bodo Rosenhahn}, \bibinfo{person}{Frank
  Hutter}, {and} \bibinfo{person}{Marius Lindauer}.}
  \bibinfo{year}{2021}\natexlab{}.
\newblock \showarticletitle{CARL: A benchmark for contextual and adaptive
  reinforcement learning}.
\newblock \bibinfo{journal}{\emph{arXiv preprint arXiv:2110.02102}}
  (\bibinfo{year}{2021}).
\newblock


\bibitem[Berger(2013)]%
        {berger2013statistical}
\bibfield{author}{\bibinfo{person}{James~O Berger}.}
  \bibinfo{year}{2013}\natexlab{}.
\newblock \bibinfo{booktitle}{\emph{Statistical decision theory and Bayesian
  analysis}}.
\newblock \bibinfo{publisher}{Springer Science \& Business Media}.
\newblock


\bibitem[Burnham and Overton(1978)]%
        {burnham1978estimation}
\bibfield{author}{\bibinfo{person}{Kenneth~P Burnham} {and}
  \bibinfo{person}{Walter~Scott Overton}.} \bibinfo{year}{1978}\natexlab{}.
\newblock \showarticletitle{Estimation of the size of a closed population when
  capture probabilities vary among animals}.
\newblock \bibinfo{journal}{\emph{Biometrika}} \bibinfo{volume}{65},
  \bibinfo{number}{3} (\bibinfo{year}{1978}), \bibinfo{pages}{625--633}.
\newblock


\bibitem[Casella and Berger(2021)]%
        {casella2021statistical}
\bibfield{author}{\bibinfo{person}{George Casella} {and}
  \bibinfo{person}{Roger~L Berger}.} \bibinfo{year}{2021}\natexlab{}.
\newblock \bibinfo{booktitle}{\emph{Statistical inference}}.
\newblock \bibinfo{publisher}{Cengage Learning}.
\newblock


\bibitem[Chaloner and Verdinelli(1995)]%
        {chaloner1995bayesian}
\bibfield{author}{\bibinfo{person}{Kathryn Chaloner} {and}
  \bibinfo{person}{Isabella Verdinelli}.} \bibinfo{year}{1995}\natexlab{}.
\newblock \showarticletitle{Bayesian experimental design: A review}.
\newblock \bibinfo{journal}{\emph{Statistical science}} (\bibinfo{year}{1995}),
  \bibinfo{pages}{273--304}.
\newblock


\bibitem[Demets and Lan(1994)]%
        {demets1994interim}
\bibfield{author}{\bibinfo{person}{David~L Demets} {and}
  \bibinfo{person}{KK~Gordon Lan}.} \bibinfo{year}{1994}\natexlab{}.
\newblock \showarticletitle{Interim analysis: the alpha spending function
  approach}.
\newblock \bibinfo{journal}{\emph{Statistics in medicine}}
  \bibinfo{volume}{13}, \bibinfo{number}{13-14} (\bibinfo{year}{1994}),
  \bibinfo{pages}{1341--1352}.
\newblock


\bibitem[{DeMets, David }(1 17)]%
        {ld98}
\bibfield{author}{\bibinfo{person}{{DeMets, David }}.}
  \bibinfo{year}{2003-11-17}\natexlab{}.
\newblock \bibinfo{booktitle}{\emph{Programs for Computing Group Sequential
  Boundaries Using the Lan-DeMets Method,}}.
\newblock
\newblock
\shownote{\url{https://biostat.wiscweb.wisc.edu/resources/software/}}.


\bibitem[Deng(2015)]%
        {deng2015objective}
\bibfield{author}{\bibinfo{person}{Alex Deng}.}
  \bibinfo{year}{2015}\natexlab{}.
\newblock \showarticletitle{Objective bayesian two sample hypothesis testing
  for online controlled experiments}. In \bibinfo{booktitle}{\emph{Proceedings
  of the 24th International Conference on World Wide Web}}.
  \bibinfo{pages}{923--928}.
\newblock


\bibitem[Deng et~al\mbox{.}(2016)]%
        {deng2016continuous}
\bibfield{author}{\bibinfo{person}{Alex Deng}, \bibinfo{person}{Jiannan Lu},
  {and} \bibinfo{person}{Shouyuan Chen}.} \bibinfo{year}{2016}\natexlab{}.
\newblock \showarticletitle{Continuous monitoring of A/B tests without pain:
  Optional stopping in Bayesian testing}. In \bibinfo{booktitle}{\emph{2016
  IEEE international conference on data science and advanced analytics
  (DSAA)}}. IEEE, \bibinfo{pages}{243--252}.
\newblock


\bibitem[Dienes(2011)]%
        {dienes2011bayesian}
\bibfield{author}{\bibinfo{person}{Zoltan Dienes}.}
  \bibinfo{year}{2011}\natexlab{}.
\newblock \showarticletitle{Bayesian versus orthodox statistics: Which side are
  you on?}
\newblock \bibinfo{journal}{\emph{Perspectives on Psychological Science}}
  \bibinfo{volume}{6}, \bibinfo{number}{3} (\bibinfo{year}{2011}),
  \bibinfo{pages}{274--290}.
\newblock


\bibitem[Foster et~al\mbox{.}(2021)]%
        {foster2021deep}
\bibfield{author}{\bibinfo{person}{Adam Foster}, \bibinfo{person}{Desi~R
  Ivanova}, \bibinfo{person}{Ilyas Malik}, {and} \bibinfo{person}{Tom
  Rainforth}.} \bibinfo{year}{2021}\natexlab{}.
\newblock \showarticletitle{Deep adaptive design: Amortizing sequential
  bayesian experimental design}. In \bibinfo{booktitle}{\emph{International
  Conference on Machine Learning}}. PMLR, \bibinfo{pages}{3384--3395}.
\newblock


\bibitem[Gelman et~al\mbox{.}(1995)]%
        {gelman1995bayesian}
\bibfield{author}{\bibinfo{person}{Andrew Gelman}, \bibinfo{person}{John~B
  Carlin}, \bibinfo{person}{Hal~S Stern}, {and} \bibinfo{person}{Donald~B
  Rubin}.} \bibinfo{year}{1995}\natexlab{}.
\newblock \bibinfo{booktitle}{\emph{Bayesian data analysis}}.
\newblock \bibinfo{publisher}{Chapman and Hall/CRC}.
\newblock


\bibitem[Gordon~Lan and DeMets(1983)]%
        {gordon1983discrete}
\bibfield{author}{\bibinfo{person}{KK Gordon~Lan} {and}
  \bibinfo{person}{David~L DeMets}.} \bibinfo{year}{1983}\natexlab{}.
\newblock \showarticletitle{Discrete sequential boundaries for clinical
  trials}.
\newblock \bibinfo{journal}{\emph{Biometrika}} \bibinfo{volume}{70},
  \bibinfo{number}{3} (\bibinfo{year}{1983}), \bibinfo{pages}{659--663}.
\newblock


\bibitem[Gupta et~al\mbox{.}(2019)]%
        {gupta2019top}
\bibfield{author}{\bibinfo{person}{Somit Gupta}, \bibinfo{person}{Ronny
  Kohavi}, \bibinfo{person}{Diane Tang}, \bibinfo{person}{Ya Xu},
  \bibinfo{person}{Reid Andersen}, \bibinfo{person}{Eytan Bakshy},
  \bibinfo{person}{Niall Cardin}, \bibinfo{person}{Sumita Chandran},
  \bibinfo{person}{Nanyu Chen}, \bibinfo{person}{Dominic Coey},
  {et~al\mbox{.}}} \bibinfo{year}{2019}\natexlab{}.
\newblock \showarticletitle{Top challenges from the first practical online
  controlled experiments summit}.
\newblock \bibinfo{journal}{\emph{ACM SIGKDD Explorations Newsletter}}
  \bibinfo{volume}{21}, \bibinfo{number}{1} (\bibinfo{year}{2019}),
  \bibinfo{pages}{20--35}.
\newblock


\bibitem[Haarnoja et~al\mbox{.}(2018)]%
        {haarnoja2018soft}
\bibfield{author}{\bibinfo{person}{Tuomas Haarnoja}, \bibinfo{person}{Aurick
  Zhou}, \bibinfo{person}{Pieter Abbeel}, {and} \bibinfo{person}{Sergey
  Levine}.} \bibinfo{year}{2018}\natexlab{}.
\newblock \showarticletitle{Soft actor-critic: Off-policy maximum entropy deep
  reinforcement learning with a stochastic actor}. In
  \bibinfo{booktitle}{\emph{International conference on machine learning}}.
  PMLR, \bibinfo{pages}{1861--1870}.
\newblock


\bibitem[Hallak et~al\mbox{.}(2015)]%
        {hallak2015contextual}
\bibfield{author}{\bibinfo{person}{Assaf Hallak}, \bibinfo{person}{Dotan
  Di~Castro}, {and} \bibinfo{person}{Shie Mannor}.}
  \bibinfo{year}{2015}\natexlab{}.
\newblock \showarticletitle{Contextual markov decision processes}.
\newblock \bibinfo{journal}{\emph{arXiv preprint arXiv:1502.02259}}
  (\bibinfo{year}{2015}).
\newblock


\bibitem[Huan and Marzouk(2016)]%
        {huan2016sequential}
\bibfield{author}{\bibinfo{person}{Xun Huan} {and} \bibinfo{person}{Youssef~M
  Marzouk}.} \bibinfo{year}{2016}\natexlab{}.
\newblock \showarticletitle{Sequential Bayesian optimal experimental design via
  approximate dynamic programming}.
\newblock \bibinfo{journal}{\emph{arXiv preprint arXiv:1604.08320}}
  (\bibinfo{year}{2016}).
\newblock


\bibitem[Ivanova et~al\mbox{.}(2021)]%
        {ivanova2021implicit}
\bibfield{author}{\bibinfo{person}{Desi~R Ivanova}, \bibinfo{person}{Adam
  Foster}, \bibinfo{person}{Steven Kleinegesse}, \bibinfo{person}{Michael~U
  Gutmann}, {and} \bibinfo{person}{Thomas Rainforth}.}
  \bibinfo{year}{2021}\natexlab{}.
\newblock \showarticletitle{Implicit deep adaptive design: policy-based
  experimental design without likelihoods}.
\newblock \bibinfo{journal}{\emph{Advances in Neural Information Processing
  Systems}}  \bibinfo{volume}{34} (\bibinfo{year}{2021}),
  \bibinfo{pages}{25785--25798}.
\newblock


\bibitem[Jeffreys(1961)]%
        {jeffreys1961theory}
\bibfield{author}{\bibinfo{person}{Harold Jeffreys}.}
  \bibinfo{year}{1961}\natexlab{}.
\newblock \bibinfo{booktitle}{\emph{The theory of probability}}.
\newblock \bibinfo{publisher}{Oxford University Press}.
\newblock


\bibitem[Johari et~al\mbox{.}(2017)]%
        {johari2017peeking}
\bibfield{author}{\bibinfo{person}{Ramesh Johari}, \bibinfo{person}{Pete
  Koomen}, \bibinfo{person}{Leonid Pekelis}, {and} \bibinfo{person}{David
  Walsh}.} \bibinfo{year}{2017}\natexlab{}.
\newblock \showarticletitle{Peeking at a/b tests: Why it matters, and what to
  do about it}. In \bibinfo{booktitle}{\emph{Proceedings of the 23rd ACM SIGKDD
  International Conference on Knowledge Discovery and Data Mining}}.
  \bibinfo{pages}{1517--1525}.
\newblock


\bibitem[Johari et~al\mbox{.}(2022)]%
        {johari2022always}
\bibfield{author}{\bibinfo{person}{Ramesh Johari}, \bibinfo{person}{Pete
  Koomen}, \bibinfo{person}{Leonid Pekelis}, {and} \bibinfo{person}{David
  Walsh}.} \bibinfo{year}{2022}\natexlab{}.
\newblock \showarticletitle{Always valid inference: Continuous monitoring of
  a/b tests}.
\newblock \bibinfo{journal}{\emph{Operations Research}} \bibinfo{volume}{70},
  \bibinfo{number}{3} (\bibinfo{year}{2022}), \bibinfo{pages}{1806--1821}.
\newblock


\bibitem[Johari et~al\mbox{.}(2015)]%
        {johari2015always}
\bibfield{author}{\bibinfo{person}{Ramesh Johari}, \bibinfo{person}{Leo
  Pekelis}, {and} \bibinfo{person}{David~J Walsh}.}
  \bibinfo{year}{2015}\natexlab{}.
\newblock \showarticletitle{Always valid inference: Bringing sequential
  analysis to A/B testing}.
\newblock \bibinfo{journal}{\emph{arXiv preprint arXiv:1512.04922}}
  (\bibinfo{year}{2015}).
\newblock


\bibitem[Kadam and Bhalerao(2010)]%
        {kadam2010sample}
\bibfield{author}{\bibinfo{person}{Prashant Kadam} {and}
  \bibinfo{person}{Supriya Bhalerao}.} \bibinfo{year}{2010}\natexlab{}.
\newblock \showarticletitle{Sample size calculation}.
\newblock \bibinfo{journal}{\emph{International journal of Ayurveda research}}
  \bibinfo{volume}{1}, \bibinfo{number}{1} (\bibinfo{year}{2010}),
  \bibinfo{pages}{55}.
\newblock


\bibitem[Kass and Raftery(1995)]%
        {kass1995bayes}
\bibfield{author}{\bibinfo{person}{Robert~E Kass} {and}
  \bibinfo{person}{Adrian~E Raftery}.} \bibinfo{year}{1995}\natexlab{}.
\newblock \showarticletitle{Bayes factors}.
\newblock \bibinfo{journal}{\emph{Journal of the american statistical
  association}} \bibinfo{volume}{90}, \bibinfo{number}{430}
  (\bibinfo{year}{1995}), \bibinfo{pages}{773--795}.
\newblock


\bibitem[Kundu et~al\mbox{.}(2021)]%
        {kundu2021conditional}
\bibfield{author}{\bibinfo{person}{Madan~Gopal Kundu},
  \bibinfo{person}{Sandipan Samanta}, {and} \bibinfo{person}{Shoubhik Mondal}.}
  \bibinfo{year}{2021}\natexlab{}.
\newblock \showarticletitle{Conditional power, predictive power and probability
  of success in clinical trials with continuous, binary and time-to-event
  endpoints}.
\newblock  (\bibinfo{year}{2021}).
\newblock


\bibitem[Liang et~al\mbox{.}(2018)]%
        {liang2018rllib}
\bibfield{author}{\bibinfo{person}{Eric Liang}, \bibinfo{person}{Richard Liaw},
  \bibinfo{person}{Robert Nishihara}, \bibinfo{person}{Philipp Moritz},
  \bibinfo{person}{Roy Fox}, \bibinfo{person}{Ken Goldberg},
  \bibinfo{person}{Joseph Gonzalez}, \bibinfo{person}{Michael Jordan}, {and}
  \bibinfo{person}{Ion Stoica}.} \bibinfo{year}{2018}\natexlab{}.
\newblock \showarticletitle{RLlib: Abstractions for distributed reinforcement
  learning}. In \bibinfo{booktitle}{\emph{International Conference on Machine
  Learning}}. PMLR, \bibinfo{pages}{3053--3062}.
\newblock


\bibitem[Lin(2013)]%
        {lin2013agnostic}
\bibfield{author}{\bibinfo{person}{Winston Lin}.}
  \bibinfo{year}{2013}\natexlab{}.
\newblock \showarticletitle{Agnostic notes on regression adjustments to
  experimental data: Reexamining Freedman’s critique}.
\newblock  (\bibinfo{year}{2013}).
\newblock


\bibitem[Lindsey et~al\mbox{.}(1999)]%
        {lindsey1999models}
\bibfield{author}{\bibinfo{person}{James~K Lindsey} {et~al\mbox{.}}}
  \bibinfo{year}{1999}\natexlab{}.
\newblock \showarticletitle{Models for repeated measurements}.
\newblock \bibinfo{journal}{\emph{OUP Catalogue}} (\bibinfo{year}{1999}).
\newblock


\bibitem[Maritz(2018)]%
        {maritz2018empirical}
\bibfield{author}{\bibinfo{person}{Johannes~S Maritz}.}
  \bibinfo{year}{2018}\natexlab{}.
\newblock \bibinfo{booktitle}{\emph{Empirical bayes methods}}.
\newblock \bibinfo{publisher}{Chapman and Hall/CRC}.
\newblock


\bibitem[Mnih et~al\mbox{.}(2016)]%
        {mnih2016asynchronous}
\bibfield{author}{\bibinfo{person}{Volodymyr Mnih},
  \bibinfo{person}{Adria~Puigdomenech Badia}, \bibinfo{person}{Mehdi Mirza},
  \bibinfo{person}{Alex Graves}, \bibinfo{person}{Timothy Lillicrap},
  \bibinfo{person}{Tim Harley}, \bibinfo{person}{David Silver}, {and}
  \bibinfo{person}{Koray Kavukcuoglu}.} \bibinfo{year}{2016}\natexlab{}.
\newblock \showarticletitle{Asynchronous methods for deep reinforcement
  learning}. In \bibinfo{booktitle}{\emph{International conference on machine
  learning}}. PMLR, \bibinfo{pages}{1928--1937}.
\newblock


\bibitem[Mnih et~al\mbox{.}(2015)]%
        {mnih2015human}
\bibfield{author}{\bibinfo{person}{Volodymyr Mnih}, \bibinfo{person}{Koray
  Kavukcuoglu}, \bibinfo{person}{David Silver}, \bibinfo{person}{Andrei~A
  Rusu}, \bibinfo{person}{Joel Veness}, \bibinfo{person}{Marc~G Bellemare},
  \bibinfo{person}{Alex Graves}, \bibinfo{person}{Martin Riedmiller},
  \bibinfo{person}{Andreas~K Fidjeland}, \bibinfo{person}{Georg Ostrovski},
  {et~al\mbox{.}}} \bibinfo{year}{2015}\natexlab{}.
\newblock \showarticletitle{Human-level control through deep reinforcement
  learning}.
\newblock \bibinfo{journal}{\emph{Nature}} \bibinfo{volume}{518},
  \bibinfo{number}{7540} (\bibinfo{year}{2015}), \bibinfo{pages}{529--533}.
\newblock


\bibitem[M{\"u}ller et~al\mbox{.}(2007)]%
        {muller2007simulation}
\bibfield{author}{\bibinfo{person}{Peter M{\"u}ller}, \bibinfo{person}{Don~A
  Berry}, \bibinfo{person}{Andy~P Grieve}, \bibinfo{person}{Michael Smith},
  {and} \bibinfo{person}{Michael Krams}.} \bibinfo{year}{2007}\natexlab{}.
\newblock \showarticletitle{Simulation-based sequential Bayesian design}.
\newblock \bibinfo{journal}{\emph{Journal of statistical planning and
  inference}} \bibinfo{volume}{137}, \bibinfo{number}{10}
  (\bibinfo{year}{2007}), \bibinfo{pages}{3140--3150}.
\newblock


\bibitem[Murphy et~al\mbox{.}(2001)]%
        {murphy2001marginal}
\bibfield{author}{\bibinfo{person}{Susan~A Murphy}, \bibinfo{person}{Mark~J
  van~der Laan}, \bibinfo{person}{James~M Robins}, {and}
  \bibinfo{person}{Conduct Problems Prevention~Research Group}.}
  \bibinfo{year}{2001}\natexlab{}.
\newblock \showarticletitle{Marginal mean models for dynamic regimes}.
\newblock \bibinfo{journal}{\emph{J. Amer. Statist. Assoc.}}
  \bibinfo{volume}{96}, \bibinfo{number}{456} (\bibinfo{year}{2001}),
  \bibinfo{pages}{1410--1423}.
\newblock


\bibitem[Puterman(2014)]%
        {puterman2014markov}
\bibfield{author}{\bibinfo{person}{Martin~L Puterman}.}
  \bibinfo{year}{2014}\natexlab{}.
\newblock \bibinfo{booktitle}{\emph{Markov decision processes: discrete
  stochastic dynamic programming}}.
\newblock \bibinfo{publisher}{John Wiley \& Sons}.
\newblock


\bibitem[Richardson et~al\mbox{.}(2022)]%
        {richardson2022bayesian}
\bibfield{author}{\bibinfo{person}{Thomas~S Richardson}, \bibinfo{person}{Yu
  Liu}, \bibinfo{person}{James McQueen}, {and} \bibinfo{person}{Doug Hains}.}
  \bibinfo{year}{2022}\natexlab{}.
\newblock \showarticletitle{A Bayesian Model for Online Activity Sample Sizes}.
  In \bibinfo{booktitle}{\emph{International Conference on Artificial
  Intelligence and Statistics}}. PMLR, \bibinfo{pages}{1775--1785}.
\newblock


\bibitem[Robbins(1970)]%
        {robbins1970statistical}
\bibfield{author}{\bibinfo{person}{Herbert Robbins}.}
  \bibinfo{year}{1970}\natexlab{}.
\newblock \showarticletitle{Statistical methods related to the law of the
  iterated logarithm}.
\newblock \bibinfo{journal}{\emph{The Annals of Mathematical Statistics}}
  \bibinfo{volume}{41}, \bibinfo{number}{5} (\bibinfo{year}{1970}),
  \bibinfo{pages}{1397--1409}.
\newblock


\bibitem[Sallab et~al\mbox{.}(2017)]%
        {sallab2017deep}
\bibfield{author}{\bibinfo{person}{Ahmad~EL Sallab}, \bibinfo{person}{Mohammed
  Abdou}, \bibinfo{person}{Etienne Perot}, {and} \bibinfo{person}{Senthil
  Yogamani}.} \bibinfo{year}{2017}\natexlab{}.
\newblock \showarticletitle{Deep reinforcement learning framework for
  autonomous driving}.
\newblock \bibinfo{journal}{\emph{Electronic Imaging}} \bibinfo{volume}{2017},
  \bibinfo{number}{19} (\bibinfo{year}{2017}), \bibinfo{pages}{70--76}.
\newblock


\bibitem[Sch{\"o}nbrodt et~al\mbox{.}(2017)]%
        {schonbrodt2017sequential}
\bibfield{author}{\bibinfo{person}{Felix~D Sch{\"o}nbrodt},
  \bibinfo{person}{Eric-Jan Wagenmakers}, \bibinfo{person}{Michael
  Zehetleitner}, {and} \bibinfo{person}{Marco Perugini}.}
  \bibinfo{year}{2017}\natexlab{}.
\newblock \showarticletitle{Sequential hypothesis testing with Bayes factors:
  Efficiently testing mean differences.}
\newblock \bibinfo{journal}{\emph{Psychological methods}} \bibinfo{volume}{22},
  \bibinfo{number}{2} (\bibinfo{year}{2017}), \bibinfo{pages}{322}.
\newblock


\bibitem[Schulman et~al\mbox{.}(2015)]%
        {schulman2015trust}
\bibfield{author}{\bibinfo{person}{John Schulman}, \bibinfo{person}{Sergey
  Levine}, \bibinfo{person}{Pieter Abbeel}, \bibinfo{person}{Michael Jordan},
  {and} \bibinfo{person}{Philipp Moritz}.} \bibinfo{year}{2015}\natexlab{}.
\newblock \showarticletitle{Trust region policy optimization}. In
  \bibinfo{booktitle}{\emph{International conference on machine learning}}.
  PMLR, \bibinfo{pages}{1889--1897}.
\newblock


\bibitem[Schulman et~al\mbox{.}(2017)]%
        {schulman2017proximal}
\bibfield{author}{\bibinfo{person}{John Schulman}, \bibinfo{person}{Filip
  Wolski}, \bibinfo{person}{Prafulla Dhariwal}, \bibinfo{person}{Alec Radford},
  {and} \bibinfo{person}{Oleg Klimov}.} \bibinfo{year}{2017}\natexlab{}.
\newblock \showarticletitle{Proximal policy optimization algorithms}.
\newblock \bibinfo{journal}{\emph{arXiv preprint arXiv:1707.06347}}
  (\bibinfo{year}{2017}).
\newblock


\bibitem[Shen and Huan(2021)]%
        {shen2021bayesian}
\bibfield{author}{\bibinfo{person}{Wanggang Shen} {and} \bibinfo{person}{Xun
  Huan}.} \bibinfo{year}{2021}\natexlab{}.
\newblock \showarticletitle{Bayesian sequential optimal experimental design for
  nonlinear models using policy gradient reinforcement learning}.
\newblock \bibinfo{journal}{\emph{arXiv preprint arXiv:2110.15335}}
  (\bibinfo{year}{2021}).
\newblock


\bibitem[Sodhani et~al\mbox{.}(2021)]%
        {sodhani2021multi}
\bibfield{author}{\bibinfo{person}{Shagun Sodhani}, \bibinfo{person}{Amy
  Zhang}, {and} \bibinfo{person}{Joelle Pineau}.}
  \bibinfo{year}{2021}\natexlab{}.
\newblock \showarticletitle{Multi-Task Reinforcement Learning with
  Context-based Representations}.
\newblock \bibinfo{journal}{\emph{arXiv preprint arXiv:2102.06177}}
  (\bibinfo{year}{2021}).
\newblock


\bibitem[Sutton and Barto(2018)]%
        {sutton2018reinforcement}
\bibfield{author}{\bibinfo{person}{Richard~S Sutton} {and}
  \bibinfo{person}{Andrew~G Barto}.} \bibinfo{year}{2018}\natexlab{}.
\newblock \bibinfo{booktitle}{\emph{Reinforcement learning: An introduction}}.
\newblock \bibinfo{publisher}{MIT press}.
\newblock


\bibitem[Tec et~al\mbox{.}(2022)]%
        {tec2022comparative}
\bibfield{author}{\bibinfo{person}{Mauricio Tec}, \bibinfo{person}{Yunshan
  Duan}, {and} \bibinfo{person}{Peter M{\"u}ller}.}
  \bibinfo{year}{2022}\natexlab{}.
\newblock \showarticletitle{A Comparative Tutorial of Bayesian Sequential
  Design and Reinforcement Learning}.
\newblock \bibinfo{journal}{\emph{arXiv preprint arXiv:2205.04023}}
  (\bibinfo{year}{2022}).
\newblock


\bibitem[Vallat(2018)]%
        {vallat2018pingouin}
\bibfield{author}{\bibinfo{person}{Raphael Vallat}.}
  \bibinfo{year}{2018}\natexlab{}.
\newblock \showarticletitle{Pingouin: statistics in Python.}
\newblock \bibinfo{journal}{\emph{J. Open Source Softw.}} \bibinfo{volume}{3},
  \bibinfo{number}{31} (\bibinfo{year}{2018}), \bibinfo{pages}{1026}.
\newblock


\bibitem[Van~Hasselt et~al\mbox{.}(2016)]%
        {van2016deep}
\bibfield{author}{\bibinfo{person}{Hado Van~Hasselt}, \bibinfo{person}{Arthur
  Guez}, {and} \bibinfo{person}{David Silver}.}
  \bibinfo{year}{2016}\natexlab{}.
\newblock \showarticletitle{Deep reinforcement learning with double
  q-learning}. In \bibinfo{booktitle}{\emph{Proceedings of the AAAI conference
  on artificial intelligence}}, Vol.~\bibinfo{volume}{30}.
\newblock


\bibitem[Vlassis et~al\mbox{.}(2012)]%
        {vlassis2012bayesian}
\bibfield{author}{\bibinfo{person}{Nikos Vlassis}, \bibinfo{person}{Mohammad
  Ghavamzadeh}, \bibinfo{person}{Shie Mannor}, {and} \bibinfo{person}{Pascal
  Poupart}.} \bibinfo{year}{2012}\natexlab{}.
\newblock \showarticletitle{Bayesian reinforcement learning}.
\newblock \bibinfo{journal}{\emph{Reinforcement learning}}
  (\bibinfo{year}{2012}), \bibinfo{pages}{359--386}.
\newblock


\bibitem[Wald and Wolfowitz(1948)]%
        {wald1948optimum}
\bibfield{author}{\bibinfo{person}{Abraham Wald} {and} \bibinfo{person}{Jacob
  Wolfowitz}.} \bibinfo{year}{1948}\natexlab{}.
\newblock \showarticletitle{Optimum character of the sequential probability
  ratio test}.
\newblock \bibinfo{journal}{\emph{The Annals of Mathematical Statistics}}
  (\bibinfo{year}{1948}), \bibinfo{pages}{326--339}.
\newblock


\bibitem[Wan et~al\mbox{.}(2021)]%
        {wan2021multi}
\bibfield{author}{\bibinfo{person}{Runzhe Wan}, \bibinfo{person}{Xinyu Zhang},
  {and} \bibinfo{person}{Rui Song}.} \bibinfo{year}{2021}\natexlab{}.
\newblock \showarticletitle{Multi-objective model-based reinforcement learning
  for infectious disease control}. In \bibinfo{booktitle}{\emph{Proceedings of
  the 27th ACM SIGKDD Conference on Knowledge Discovery \& Data Mining}}.
  \bibinfo{pages}{1634--1644}.
\newblock


\bibitem[Welch(1947)]%
        {welch1947generalization}
\bibfield{author}{\bibinfo{person}{Bernard~L Welch}.}
  \bibinfo{year}{1947}\natexlab{}.
\newblock \showarticletitle{The generalization of ‘STUDENT'S’problem when
  several different population varlances are involved}.
\newblock \bibinfo{journal}{\emph{Biometrika}} \bibinfo{volume}{34},
  \bibinfo{number}{1-2} (\bibinfo{year}{1947}), \bibinfo{pages}{28--35}.
\newblock


\end{thebibliography}
